\documentclass[10pt,twocolumn,letterpaper]{article}

\usepackage[pagenumbers]{iccv} % To force page numbers, e.g. for an arXiv version

\newcommand{\ours}[0]{GLEAM\xspace}
\newcommand{\oursbench}[0]{GLEAM-Bench\xspace}

\usepackage{arydshln}

\definecolor{iccvblue}{rgb}{0.21,0.49,0.74}
\usepackage[pagebackref,breaklinks,colorlinks,allcolors=iccvblue]{hyperref}

\usepackage{gensymb}
\usepackage{color}
\usepackage{url}
\usepackage{booktabs}
\usepackage{multirow}
\usepackage{caption}
\usepackage{lipsum}
\usepackage{tabularx}
\usepackage{float}
\usepackage{stfloats}
\usepackage{makecell}
\usepackage{bbding}
\usepackage{colortbl} 
\usepackage{bm}

\def\paperID{8640} % *** Enter the Paper ID here
\def\confName{ICCV}
\def\confYear{2025}

\title{GLEAM: Learning Generalizable Exploration Policy for Active Mapping \\ in Complex 3D Indoor Scenes}
\author{Xiao Chen$^{1, 2}$ \quad Tai Wang$^{2}$ \quad Quanyi Li$^{2}$ \quad Tao Huang$^{2}$ \quad Jiangmiao Pang$^{2}$ \quad Tianfan Xue$^{1}$  \vspace{0.1cm} \\
$^1$The Chinese University of Hong Kong \quad $^2$Shanghai AI Laboratory \\
\vspace{-7pt}\\
\textbf{Project Website}: \href{https://xiao-chen.tech/gleam}{xiao-chen.tech/gleam}
}

\begin{document}
% \maketitle

%%%%%%%%% ABSTRACT

\twocolumn[{
    \renewcommand\twocolumn[1][]{#1}%
    \maketitle
    \begin{center}
        \vspace{-25pt}
        \hsize=\textwidth
        % \vspace{-30pt}
        \includegraphics[width=\textwidth]{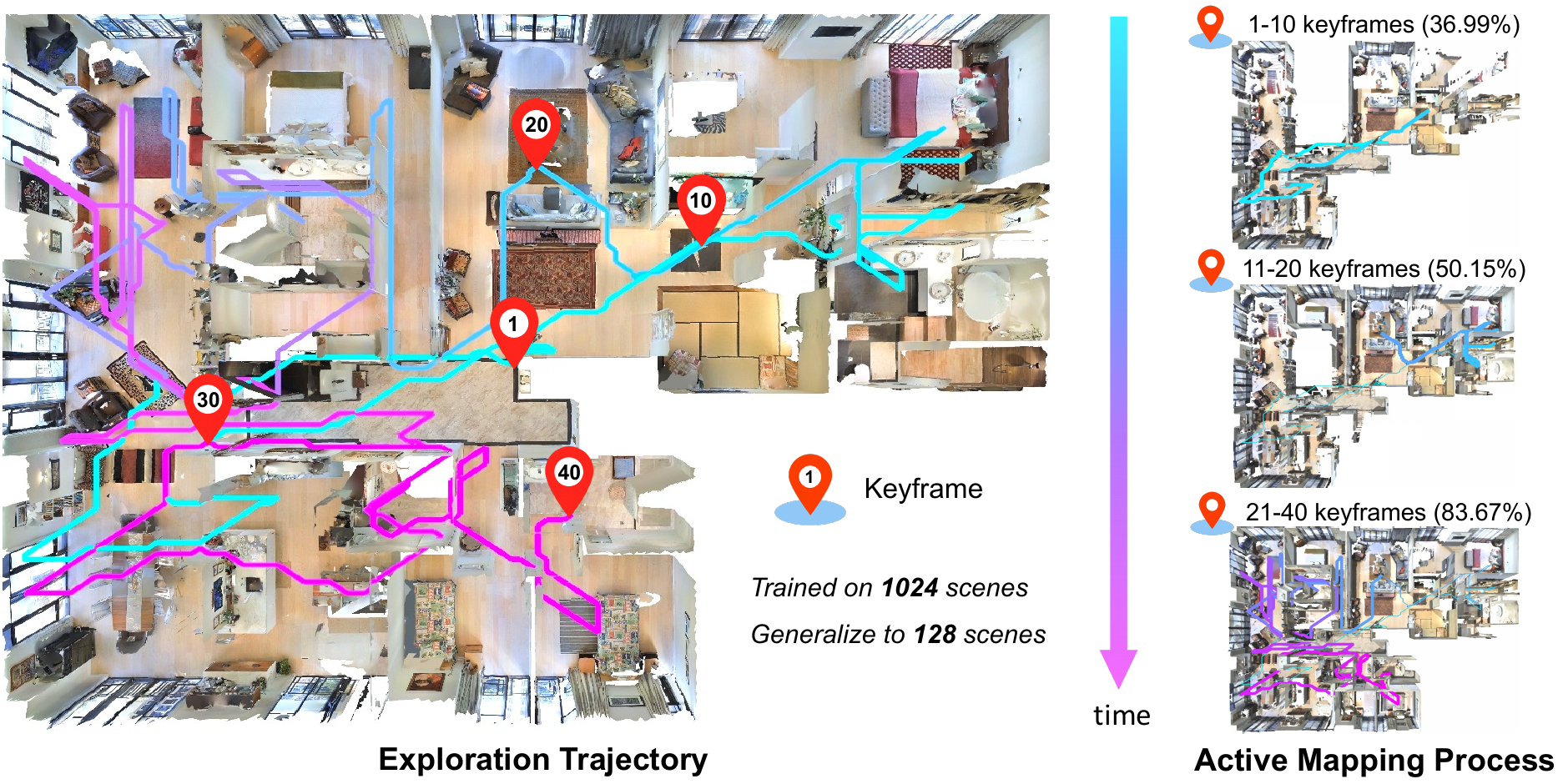}
        \captionof{figure}{We introduce \textbf{\ours}, a unified generalizable exploration policy for active mapping in complex 3D indoor scenes, trained and evaluated on 1,152 diverse scenes from our benchmark~\textbf{\oursbench}. 
        Our cross-dataset generalization to an unseen real-scan scene from Matterport3D~\cite{chang2017matterport3d} achieves 83.67\% coverage using 40 keyframes, without any fine-tuning and prior knowledge.}
        \label{fig:teaser}        
    \end{center}
}]

\begin{abstract}
    % Generalizable active mapping in complex unknown environments remains a critical challenge for mobile robots. Existing methods, constrained by insufficient training data and conservative exploration strategies, exhibit limited generalizability across scenes with diverse layouts and complex connectivity. To enable scalable training and reliable evaluation, we introduce GLEAM-Bench, the first large-scale benchmark designed for generalizable active mapping with 1,152 diverse 3D scenes from synthetic and real-scan datasets. Building upon this foundation, we propose GLEAM, a unified generalizable exploration policy for active mapping. Its superior generalizability comes mainly from our semantic representations, long-term navigable goals, and randomized strategies. It significantly outperforms state-of-the-art methods, achieving 66.50% coverage (+9.49%) with efficient trajectories and improved mapping accuracy on 128 unseen complex scenes. Project page: https://xiao-chen.tech/gleam/.
    
    Generalizable active mapping in complex unknown environments remains a critical challenge for mobile robots. 
    Existing methods, constrained by insufficient training data and conservative exploration strategies, exhibit limited generalizability across scenes with diverse layouts and complex connectivity.
    To enable scalable training and reliable evaluation, we introduce \textbf{\oursbench}, the first large-scale benchmark designed for generalizable active mapping with 1,152 diverse 3D scenes from synthetic and real-scan datasets. 
    Building upon this foundation, we propose \textbf{\ours}, a unified \underline{g}enera\underline{l}izable \underline{e}xploration policy for \underline{a}ctive \underline{m}apping.
    Its superior generalizability comes mainly from our semantic representations, long-term navigable goals, and randomized strategies.
    It significantly outperforms state-of-the-art methods, achieving 66.50\% coverage (+9.49\%) with efficient trajectories and improved mapping accuracy on 128 unseen complex scenes.
\end{abstract}
\vspace{-15pt}
\section{Introduction}
\label{sec:intro}

    While existing methods have enabled robots to represent~\cite{mildenhall2020nerf, kerbl20233d} and reconstruct~\cite{sucar2021imap, iSDF2022} 3D environments through predefined trajectories or offline visual data, autonomous exploration and mapping in unknown 3D environments remain a cornerstone challenge for robots. In unknown environments with complex connectivity, robots must strategically prioritize unexplored areas while balancing exploration efficiency. Classic active SLAM~\cite{chenlearning,chaplotlearning} and active mapping~\cite{ramakrishnan2020occupancy,georgakis2022uncertainty,yan2023anm} have been investigated in the context of small-scale or simple scenarios, often assuming in-distribution environments with few rooms.
    
    % The generalization of exploration policies for active mapping in unknown complex scenes remains inadequately studied.
    The generalization of existing active mapping methods to unknown scenes remains inadequately studied.
    Three primary challenges emerge across data, technical frameworks, and training strategies.
    First, most existing methods~\cite{chenlearning,chaplotlearning,ramakrishnan2020occupancy,georgakis2022uncertainty} are trained on fewer than 100 homogeneous scenarios, failing to leverage data diversity for robust generalization, as shown in \cref{tab:related_work}. 
    Second, many existing approaches rely on empirically defined heuristics to guide the exploration, including information gain~\cite{yan2023anm}, gain or layout anticipations~\cite{ramakrishnan2020occupancy,georgakis2022uncertainty,li2025nextbestpath}, and structured map~\cite{chenlearning,cao2024deep}. The heuristic metrics hinder their generalization in heterogeneous environments with diverse obstacle layouts and topological connectivity.
    Third, previous training settings such as centralizing starting positions~\cite{chaplotlearning} simplify the exploration pattern, thus diminishing the policy's capacity to explore complex interconnected spaces. 
    % These limitations collectively constrain the generalizability of active mapping.

\begin{figure}[!t]
        % \hsize=\textwidth
        \centering
        % \vspace{-40pt}
        % \includegraphics[width=\textwidth,height=5cm]{example-image}
        % \includegraphics[height=5cm]{example-image}
        \includegraphics[width=0.47\textwidth]{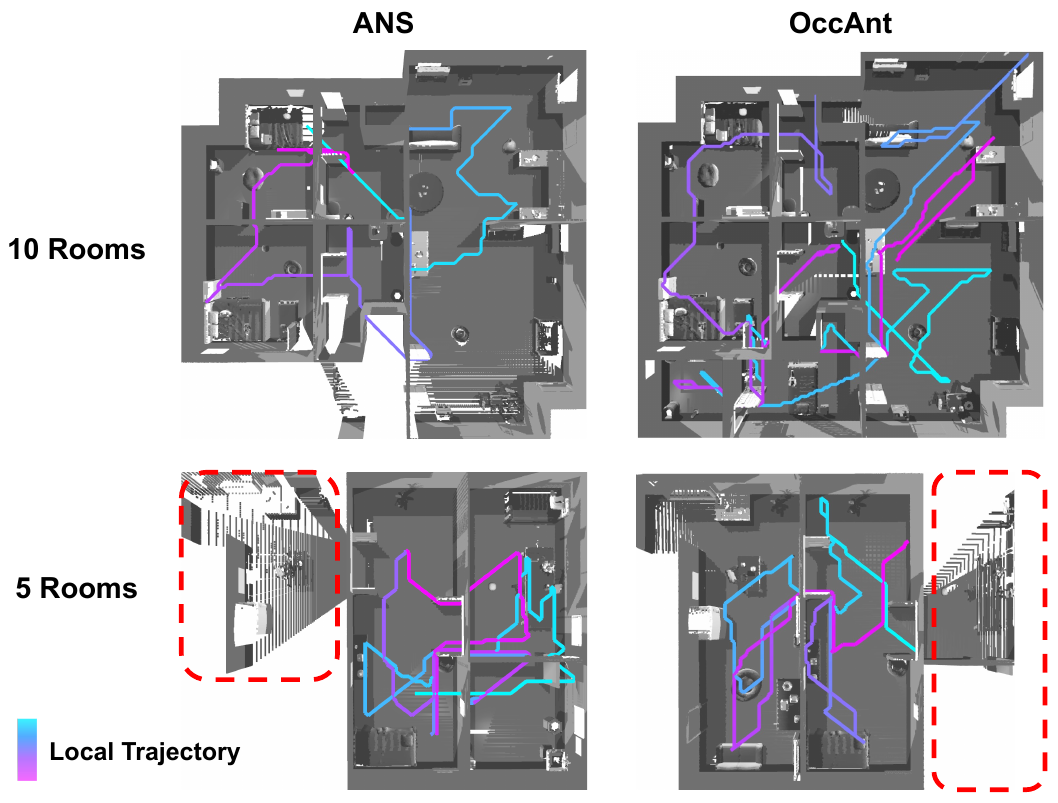}
        % \caption{Even trained on 32 complex scenes each with 10 rooms, ANS and OccAnt cannot well generalize to the heterogeneous scenes with only 5 rooms, showing the limitation of generalization.}
        \caption{Despite being trained on complex indoor scenes with 10 rooms, classic RL-based methods, ANS~\cite{chenlearning} and OccAnt~\cite{ramakrishnan2020occupancy} exhibit limited generalizability when tested on simple but structurally distinct scenes with 5 rooms.}
        \label{fig:motivation}        
    \end{figure}

    % To mitigate the impact of scarce data on the generalizability of exploration policies, we introduce \textbf{\oursbench}, the first large-scale exploration benchmark encompassing over 1,000 complex 3D indoor scenes, with parallel simulation support. 
    Therefore, to build an active mapping system that can generalize to different complex indoor scenes, we build \textbf{\oursbench}, a new training dataset and evaluation benchmark. It is the first large-scale exploration dataset encompassing over 1,000 complex 3D indoor scenes for training and more than 100 scenes for evaluation, with parallel simulation support. To show the importance of using a large-scale benchmark, we train the previous state-of-the-art active mapping algorithms, ANS~\cite{chaplotlearning} and OccAnt~\cite{ramakrishnan2020occupancy}, on 32 scenes with 12 rooms. As shown in \cref{fig:motivation}, the trained models can perform well in scenes with 10 rooms, but they fail in even simpler environments with only 5 rooms, demonstrating their poor generalization. Therefore, in order to ensure the generalization of new exploration policies, both the training and evaluation datasets contain indoor scenes from different datasets, including both synthetic ones (ProcTHOR~\cite{deitke2022procthor}, HSSD~\cite{khanna2024hssd}) and real-scanned ones (Gibson~\cite{xia2018gibson}, Matterport3D~\cite{chang2017matterport3d}), with 1,152 scenes in total, ensuring both diversity and complexity.
    % To validate the effect of training scenes, we train and ablate our exploration policy on 1,024 diverse scenes from synthetic (ProcTHOR, HSSD) and real-scan (Gibson) datasets, revealing the contribution of quantity, diversity, and complexity of training scenes.

    This new benchmark motivates a better active mapping algorithm that can generalize across diversified indoor scenes. Therefore, we subsequently propose \textbf{\ours}, a reinforcement learning (RL)-based exploration policy that can be generalized to complex unseen indoor scenes without any fine-tuning or prior knowledge. Our policy achieves better generalization through the following three key designs. First, it maintains a global probabilistic map that integrates historical observations and a semantic egocentric map with four task-related states, enabling environment-agnostic spatial reasoning—the lightweight LocoTransformer~\cite{yang2021learning} distills these states into task-aware embeddings without scene-specific architectural constraints. Second, we replace classic motion primitives with long-horizon action spaces validated by heuristic planners, enabling agents to focus on high-level exploration while offloading low-level path safety at the early training stage. Third, the training strategies include randomized initial poses to enhance generalization by exposing agents to broader environmental variations, thereby mitigating scene-specific overfitting and improving policy robustness.

    At last, to validate the generalizability of \ours, we conduct comprehensive evaluations on our benchmark comprising 128 challenging scenes from four distinct datasets. Without any fine-tuning, \ours achieves 66.50\% average coverage ratio across all scenes, surpassing the previous state-of-the-art ANM~\cite{yan2023anm} by 11.41\% while concurrently improving path efficiency (+9.49\% AUC of coverage) and reconstruction completeness (-0.22m nearest distance).

    \begin{table}[!t]
        \centering
        %\small
        % \setlength\tabcolsep{6pt}  % width of column
        % \renewcommand{\arraystretch}{1.0}   % height of row
        % \captionof{table}{The sources of our collected and processed scene meshes. We created the meshes of complex scenes from ProcTHOR-10K using our export script. After manually filtering out the meshes of complex indoor scenes from \textbf{ProcTHOR-10K}, \textbf{HSSD}, \textbf{Gibson}, and \textbf{MP3D}, we preprocess them and split them into \textbf{1024 training scenes} and \textbf{128 test scenes}.}
        \captionof{table}{The data sources of existing exploration policies for active mapping. $\dagger$: The SUN-CG dataset is not available now. ``*": The policies are based on neural representations and thus need per-scene optimization on evaluation scenes.}
        \renewcommand{\arraystretch}{1.1}   % height of row
    \resizebox{1\linewidth}{!}{
        \begin{tabular}{llcc}
        \toprule
        \textbf{Method} & \textbf{Data Source} & \textbf{Training} & \textbf{Evaluation} \\ \midrule
        ExploreNav~\cite{chenlearning} & SUN-CG$\dagger$~\cite{song2017semantic} & 20 & 20 \\
        ANS~\cite{chaplotlearning} & SUN-CG$\dagger$ & 20 & 20  \\
        OccAnt~\cite{ramakrishnan2020occupancy} & Gibson~\cite{xia2018gibson}, MP3D~\cite{chang2017matterport3d} & 72 & 18 \\
        UPEN~\cite{georgakis2022uncertainty} & MP3D & 61 & 18  \\
        ANM*~\cite{yan2023anm} & Gibson, MP3D & N/A & 14 \\
        NARUTO*~\cite{feng2024naruto} & Replica~\cite{straub2019replica}, MP3D & N/A & 13 \\
        % \multirow{2}{*}{\ours} & ProcTHOR, HSSD, & \multirow{2}{*}{\textbf{1024}} & \multirow{2}{*}{\textbf{128}} \\
        %  & Gibson, MP3D & & \\
        \midrule
        \ours & ProcTHOR~\cite{deitke2022procthor}, MP3D, & \multirow{2}{*}{\textbf{1,024}} & \multirow{2}{*}{\textbf{128}} \\
        (Ours) & HSSD~\cite{khanna2024hssd}, Gibson & & \\
        \bottomrule
        \end{tabular}}
        \vspace{-5pt}
        \label{tab:related_work}
    \end{table}

\section{Related Work}
\label{sec:related_work}
\vspace{-1pt}

\noindent\textbf{Active Mapping.}
    Active mapping is a promising field yet to be thoroughly benchmarked. %, primarily because of the different task goals, sensor platforms, and exploration targets.
    The active mapping pipeline alternates between inferring the next optimal viewpoints, capturing new data, and updating the built 3D modeling. Classic active SLAM~\cite{lluvia2021active, chenlearning,chaplotlearning} and active mapping~\cite{ramakrishnan2020occupancy,georgakis2022uncertainty,yan2023anm} have been investigated in the context of small-scale or simple scenarios, often assuming in-distribution environments with few rooms. 
    % In unknown environments with complex internal connectivity, robots must strategically prioritize unexplored areas while balancing exploration efficiency. 
    % Typical exploration policies in active 3D mapping systems can be roughly categorized into: 1) heuristic, 2) information gain-based, and 3) reinforcement learning-based policies.
    As one of the most representative heuristic policies, frontier-based exploration (FBE) policy~\cite{Yamauchi1997FBE, dornhege2013frontier, batinovic2021multi} recognizes the boundary between explored areas and unknown areas and then navigates agents to the frontier. However, these policies usually rely on impractical criteria such as always moving to the closest frontier and don't leverage the semantic priors like diverse indoor layouts, thus cannot effectively generalize to the complex unseen scenes in the real world. % As for frontier selection criteria, classic policies~\cite{Yamauchi1997FBE, dornhege2013frontier, batinovic2021multi} always choose the closest frontier as the next navigation target, which is proven suboptimal in complex scenes. To resolve it, some works~\cite{dai2020fast}~\xiao{need more} incorporate next-best-frontier selection criteria into the exploration policies. However, these policies suffer from long-term exploration inefficiency due to the nature of their step-by-step greedy decision process.

    Information gain-based policies are also known as uncertainty-driven or utility-driven policies.
    % The typical frameworks consist of an information gain estimator and a next-best-view selector. 
    Classic works~\cite{isler2016information, bircher2016receding} use information theory to quantify the information gain for a probabilistic volumetric representation. A recent group of works~\cite{lee2022uncertainty, zhan2022activermap, ran2023neurar, yan2023anm,jiang2023fisherrf} leverages neural implicit representation, such as SDF~\cite{park2019deepsdf} and NeRF~\cite{mildenhall2020nerf}, to model the uncertainty fields of target scenes.

\noindent\textbf{Existing benchmark for Exploration.}
Existing benchmarks~\cite{straub2019replica,xia2018gibson,chang2017matterport3d,dai2017scannet,wang2024embodiedscan,yeshwanth2023scannet++} largely derived from 3D datasets tailored for perception, short-range navigation, or small-scale scene reconstruction. These benchmarks suffer from three critical constraints: (1) Restricted environmental scales with discontinuous room layouts, inadequate for long-horizon task requirements; (2) Oversimplify collision dynamics by neglecting dense obstacle arrangements, resulting in policies that struggle in cluttered real-world settings; (3) Fragmented geometric surfaces in real-scan environments that compromise exploration policies' ability to apply learned scene priors.
% First, their scenes are often limited to small-scale environments, lacking the multi-room continuity necessary for long-horizon tasks. Second, while these datasets emphasize visual diversity, they oversimplify collision dynamics by neglecting dense obstacle arrangements, resulting in policies that struggle in cluttered real-world settings. Third, real-scan scenes' fragmented surface geometries confuse exploration policies in utilizing learned scene priors. 
While these benchmarks enable basic validation of exploration efficiency and mapping quality, they lack diversity and complexity in scene layouts, asset quality, and topological structures. 
% As shown in Fig.~\ref{fig:motivation}, methods like ANS~\cite{chaplotlearning} and OccAnt~\cite{ramakrishnan2020occupancy} are trained on datasets with fewer than 32 homogeneous scenes, limiting their ability to generalize to environments with varying scales or structural heterogeneity. This scarcity of large-scale, diverse benchmarks restricts the development of active mapping algorithms that can adapt to real-world complexities.

    \begin{figure*}[!t]
        \vspace{-10pt}
      \begin{minipage}[b]{0.7\textwidth}
            \centering
            % \setlength\tabcolsep{6pt}  % width of column
            % \small
            \normalsize
            \renewcommand{\arraystretch}{1.0}   % height of row              
            \begin{tabular}{lcccccc}
            \hline
            \textbf{Dataset} & \textbf{Type} & \textbf{\#Scenes} & \textbf{NS($m^2$)} & \textbf{FS($m^2$)} & \textbf{NC} & \textbf{SC} \\ \hline
            Replica~\cite{straub2019replica} & synthetic & 18 & 0.56k & 2.19k & 5.99 & 3.40 \\
            HSSD~\cite{khanna2024hssd} & synthetic & 211 & 53.21k & - & 13.7 & 5.90 \\
            ProcTHOR-10K~\cite{deitke2022procthor} & synthetic & 10k & 220k & - & - & - \\
            Gibson (4+ only)~\cite{xia2018gibson} & real-scan & 106 & 7.18k & 17.74k & 11.90 & 3.04 \\
            Matterport3D~\cite{chang2017matterport3d} & real-scan & 90 & 30.22k & 101.82k & 17.09 & 2.99 \\
            \rowcolor{gray!20}
            \oursbench & mixed & 1,152 & 91.16k & 164.66k & 11.35 & 3.44\\
            \hline
            \end{tabular}
            % \vspace{5pt}
            \vspace{-6pt}
            \captionof{table}{The description of the mentioned dataset in Table 1. Note that only a few high-quality scenes from these datasets are used by previous active mapping algorithms. \textbf{NS}: navigable space, \textbf{FS}: floor space, \textbf{NC}: navigation complexity, \textbf{SC}: scene clutter. ``-": challenging to access without annotation or statistical documentation.}
            % ``*": ProcTHOR-10K doesn't provide meshes, thus we estimate its total navigable space according to the provided information.}      
            \label{tab:benchmark}            
      \end{minipage}
      \hfill
      \begin{minipage}[b]{0.25\textwidth}
            \centering            
            \includegraphics[width=0.9\linewidth, height=4cm]{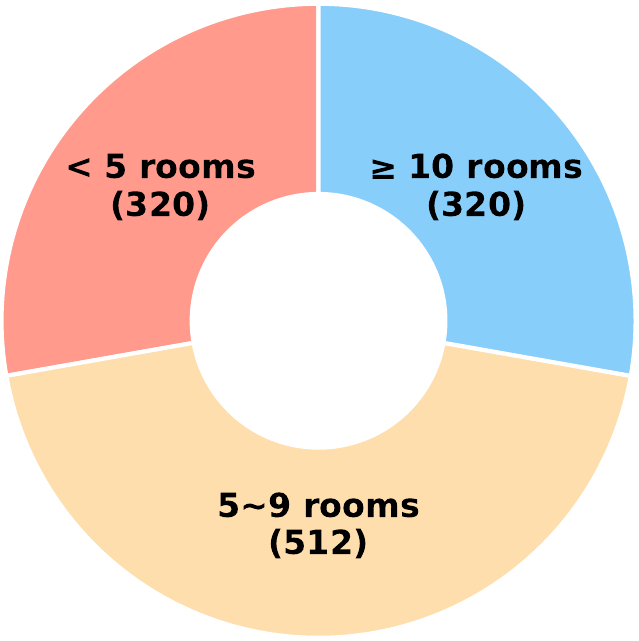} 
            \captionof{figure}{The distribution of 1,152 scenes by the number of rooms in our benchmark \oursbench.}
            \vspace{5pt}
            \label{fig:benchmark}
      \end{minipage}
      % \vspace{-12pt}
    \end{figure*}

\noindent\textbf{The Generalizability of Exploration Policies.}
The generalization of exploration policies for active mapping in unknown environments remains underexplored, with three key limitations in existing approaches. First, data scarcity and homogeneity hinder robust generalization: most methods~\cite{chenlearning,chaplotlearning,ramakrishnan2020occupancy,georgakis2022uncertainty} are trained on fewer than 100 scenes, often with uniform room counts or layouts, leading to overfitting specific or simple environmental patterns. Second, classic technical frameworks rely on rigid heuristics, such as handcrafted information gain~\cite{yan2023anm}, layout anticipations~\cite{ramakrishnan2020occupancy,li2025nextbestpath}, or structured map priors~\cite{chenlearning,cao2024deep}, which struggle to adapt to unseen obstacle configurations or topological variations. For example, methods using motion primitives or fixed action spaces~\cite{ramakrishnan2020occupancy,chenlearning} fail to handle long-horizon exploration in interconnected spaces. Third, simplified training strategies—such as centralized agent starting positions~\cite{chenlearning,ramakrishnan2020occupancy}—reduce exposure to environmental variability, weakening policies’ adaptability to complex initial conditions. These limitations collectively constrain the deployment of exploration policies in real-world scenarios with heterogeneous layouts and dynamic connectivity.

\vspace{-3pt}
\section{\oursbench}
\label{sec:bench}
We introduce \oursbench, a benchmark for generalizable exploration for active mapping in complex 3D indoor scenes. The statistical metrics can be found in Table~\ref{tab:benchmark} and Fig.~\ref{fig:benchmark}, following~\cite{ramakrishnan2021hm3d,chang2017matterport3d,wang2024embodiedscan}.
These scene meshes are characterized by watertight geometry, diverse floorplan ($\geq$10 types), and complex interconnectivity. We unify and refine multi-source datasets through manual filtering, geometric repair, and task-oriented preprocessing. 
To simulate the exploration process, we connect our dataset with NVIDIA Isaac Gym~\cite{makoviychuk2021isaac}, enabling parallel sensory data simulation and online policy training, achieving 150 FPS on an RTX 3090 GPU, even trained on 512 complex scenes. Additional details about \oursbench are provided in Appendix~\ref{app:data}.

% To address these shortcomings. Our benchmark features \textbf{\textit{1152} diverse complex 3D scenes} from different indoor datasets, including realistic synthetic datasets (ProcTHOR~\cite{deitke2022procthor}, HSSD~\cite{khanna2024hssd}) and two real-scan datasets (Gibson~\cite{xia2018gibson}, Matterport3D~\cite{chang2017matterport3d}). We create, filter, and preprocess them by the criteria of active mapping. \oursbench redefines the standards for exploration-oriented benchmarks. Our scenes are meticulously curated to include large-scale diverse layouts—such as 2-bed-2-bath and multi-family apartments—that mirror the complexity of real buildings. We incorporate high-density clutter (e.g., furniture, appliances) and geometrically intricate surfaces to simulate realistic obstacle distributions and contact dynamics. 

    \begin{figure*}[!t]
        \hsize=\textwidth
        \centering
        \includegraphics[width=\textwidth]{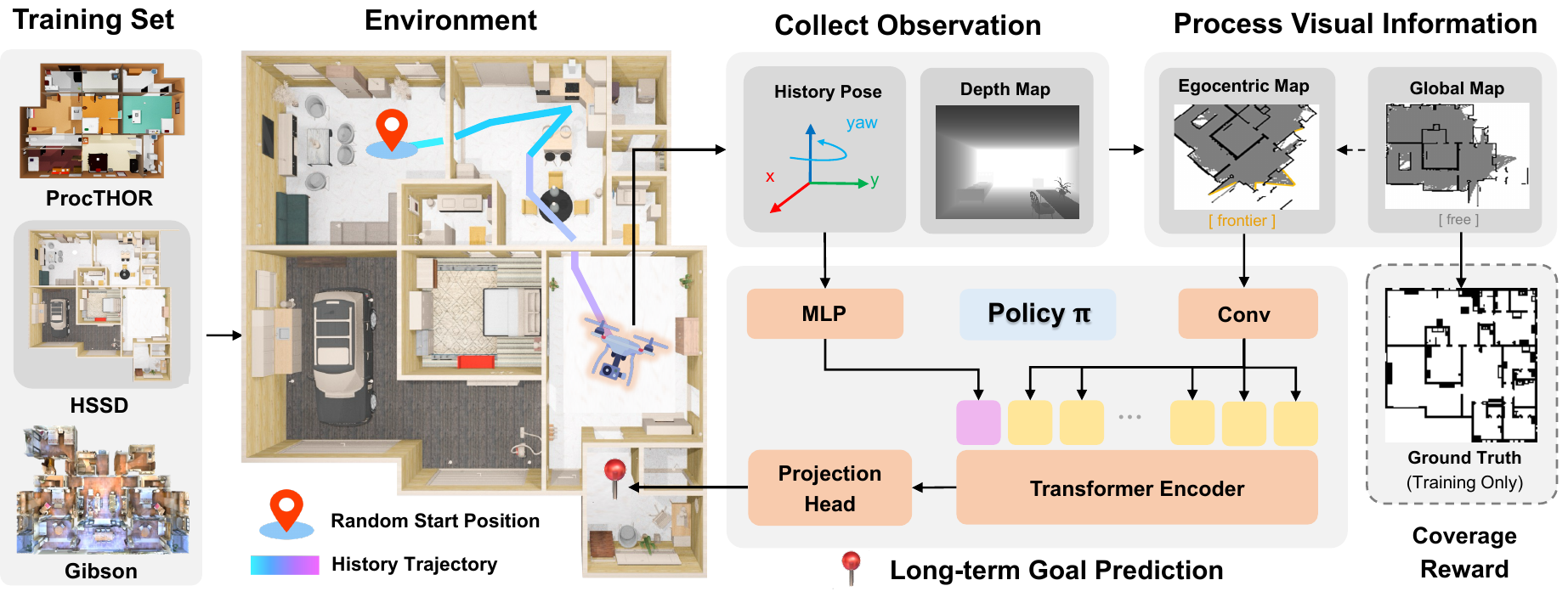}
        \caption{The overview of our framework. Trained on 1,024 diverse indoor scenes, \ours processes depth observations and agents' poses to iteratively update a global map. An egocentric map is extracted and augmented with exploration frontiers to capture semantic exploration cues. A lightweight Transformer encoder then analyzes the egocentric map and trajectory history to predict the long-term goals. The reward function of coverage is computed by the global map and ground-truth occupancy map.}
        \label{fig:overview}        
        \vspace{-3pt}
    \end{figure*}

\vspace{-1pt}
\subsection{Datasets}
    % Most existing methods are trained on fewer than 100 homogeneous scenarios, failing to leverage data diversity for robust generalization.

    % \noindent \textbf{ProcTHOR.}
    % ProcTHOR~\cite{deitke2022procthor} and AI2-THOR~\cite{kolve2022ai2thor} have empowered the research community to procedurally generate interactive, high-fidelity indoor scenes with diverse layouts for robotic training at scale. 
    Our benchmark features \textbf{\textit{1,152} diverse complex 3D scenes} from different indoor datasets, including realistic synthetic datasets (ProcTHOR~\cite{deitke2022procthor}, HSSD~\cite{khanna2024hssd}) and two real-scan datasets (Gibson~\cite{xia2018gibson}, Matterport3D~\cite{chang2017matterport3d}). 
    It is built to overcome the problem of data scarcity and homogeneity faced by most existing methods, as shown in Table~\ref{tab:related_work}. Concretely, we scale up, filter, and preprocess scenes using unified criteria to build our dataset. For example, we only select enclosed scene meshes with a nearly watertight external surface and high-quality real-scan meshes with minimal floaters and artifacts. However, unlike other datasets providing mesh files directly, we can only access the digital assets of ProcTHOR through the Habitat platform~\cite{savva2019habitat} and Unity Engine~\cite{juliani2020unity}, as they are stored and organized in a proprietary format. To address this and extend these valuable synthetic data usages, we developed an autonomous script that can launch the generation given configuration, i.e., number of rooms, and then batch export the generated scenes from Unity Editor to mesh files. The resulting scene meshes have multiple editable contents, including materials, floorplans, object placement, and controllable connectivity. We believe these large-scale synthetic datasets will benefit many data-driven applications and will release both the export script and the created assets.

\subsection{Complex Scenes for Long-Horizon Exploration}
% Unlike prior benchmarks repurposed from perception-focused datasets, our scenes prioritize structural scalability, collision complexity, and physical fidelity to bridge critical gaps in training exploration policies.
Largely derived from 3D datasets tailored for perception or short-range navigation, previous long-horizon exploration benchmarks~\cite{straub2019replica,xia2018gibson} exhibit limitations such as fragmented surfaces and simple layouts, detailed in Section~\ref{sec:related_work}. 
% First, their scenes are often limited to small-scale environments, lacking the multi-room continuity essential for long-horizon tasks. Second, while these datasets emphasize visual diversity, they oversimplify collision dynamics by neglecting dense obstacle arrangements, resulting in policies that struggle in cluttered real-world settings. Third, the surface geometries of real-scan scenes are spatially fragmented, confusing the exploration policy in utilizing the learned scene prior. 
% In brief, the impractical and low-quality scenes undermine the sim-to-real generalizability of exploration policies.
% each featuring eight or more interconnected rooms and architecturally realistic layouts specifically designed for long-horizon exploration. 
% \oursbench redefines the standards for exploration-oriented benchmarks. Our
As a part of \oursbench, approximately 400 complex indoor scenes aim to address these shortcomings. They are meticulously curated to include large-scale, diverse but architecturally realistic layouts, such as 2-bed-2-bath and multi-family apartments that mirror the complexity of real buildings. We incorporate high-density clutter (e.g., furniture, appliances) and geometrically intricate surfaces to simulate realistic obstacle distributions and contact dynamics.

% Furthermore, leveraging these meshes collected for exploration, we achieve millimeter-level geometric accuracy and physically plausible interaction properties, ensuring that policies trained in our benchmark generalize to unstructured environments.

% Central to our design is a focus on long-horizon exploration challenges. Scenes enforce realistic multi-room connectivity, requiring agents to reason over sequential navigation, backtracking, and dynamic occlusion patterns caused by strategically placed objects. To emulate real-world unpredictability, we incorporate configurable object placements and lighting variations, compelling policies to adapt to non-stationary conditions. By integrating these features, \oursbench not only advances the methodological rigor of exploration research but also provides a scalable foundation for deploying robust, real-world exploration strategies.

\section{Methodology}
\label{sec:method}

\subsection{Task Formulation}
\label{sec:method_form}
    \noindent We formulate the active mapping problem as learning an optimal exploration policy $\pi$ that controls an agent to capture enough information of unseen target scenes for 3D reconstruction, with limited decision-making budgets and safety constraints. 
    Following~\cite{chenlearning, ramakrishnan2020occupancy, chen2024gennbv}, we adopt a reinforcement learning (RL)-based framework to model these long-horizon exploration tasks for active mapping. % In our setting, the agent alternates between capturing observations, updating current maps, and navigating to a predicted long-term goal. 

    As shown in Fig.~\ref{fig:overview}, our simulated agent is embodied as CrazyFlie~\cite{giernacki2017crazyflie}, a type of unmanned aerial vehicle equipped with an onboard depth camera and an IMU, to execute data collection for reconstruction. Our agent's observation $o_t$ includes the depth map and its poses at each time step $t$. After capturing novel observations, we update the maintained egocentric map $M_t$ (Sec.~\ref{sec:method_map}) as the agent's current state $s_t$. Based on the map, our RL-based policy predicts the action $a_t$ (Sec.~\ref{sec:method_action}) including movement and orientation to the navigable long-term goal. To learn a generalizable unified exploration policy, we propose effective training strategies (Sec.~\ref{sec:method_strategy}) and optimization settings (Sec.~\ref{sec:method_opt}).

    % We expect our unified exploration policy can be generalized to unseen scenes without any fine-tuning, achieve efficient exploration trajectories, and obtain accurate reconstruction results.

% \subsection{Egocentric Semantic Map}
% \subsection{Semantic Map Representations~\xiao{Title}}
\subsection{Semantic Map Representations}
\label{sec:method_map}

    As Fig.~\ref{fig:overview} shows, we maintain two maps during exploration: 1) an egocentric semantic map $M_t$ as the agent's state $s_t$, and 2) a probabilistic global map $G_t$ in the world coordinate, designed to integrate historical observations. 
    Each cell of the semantic egocentric map $M_t$ is categorized as task-related states among {occupied, free, unknown, frontier}, enabling environment-agnostic spatial reasoning.

    We begin by detailing how to construct and update $G_t$. At the initial time step $t=0$, the depth map $D_0$ is back-projected into a 3D point cloud in the world coordinate using the camera's intrinsic and extrinsic parameters. The raw point cloud is filtered to retain only key points within a predefined height range. These filtered points are then projected into a temporary binary occupancy map via top-down projection. Inspired by~\cite{chen2024gennbv}, we extend the binary occupancy map to the probabilistic occupancy map $G_t$ that distinguishes the unexplored area and free area to guide the exploration process.
    To construct the probabilistic map, we adopt Bresenham's Line Algorithm~\cite{bresenham1965algorithm} to cast the ray path in this temporary binary map from the agent's position to the endpoints among occupied cells. Following the classical occupancy grid mapping algorithm~\cite{thrun2002probabilistic,chen2024gennbv}, we have the log-odds formulation of occupancy probability:
        \begin{equation}
        \log \mathrm{Odd}(m_i|z_j) = \log \mathrm{Odd}(m_i) + C,
        \end{equation}
    where $m_i$ is the occupancy probability of $i^{th}$ cell in the map $M_t$, $z_j$ is the measurement event that $j^{th}$ camera ray passes through this cell, and $C=\log\frac{p(z_j|m_i=1)}{p(z_j|m_i=0)}$ can be regarded as an empirical constant. The derivation can be found in the Appendix. At each step $t+1$, we update the probabilistic global map $G_{t+1}$ based on the preceding map $G_t$ and newly observed points. Each cell of the global map can be discretized into three categories among \{occupied, free, unknown\}. 

    We proceed to elaborate on the construction of the egocentric semantic map $M_t$. At each time step $t$, we extract an egocentric map centered at the agent's current position from the global map $G_t$. To enhance the exploration-oriented semantics, we implement a frontier detection module inspired by~\cite{Yamauchi1997FBE} to extend the category of frontiers. Specifically, we define a convolutional kernel to recognize boundaries between free and unknown areas. The resulting egocentric map $M_t$ preserves geometric structures while encoding exploration progress, enabling effective decision-making through differentiable spatial reasoning.

    After each round of map updating,  the global map $G_t$ is compared with the ground-truth occupancy map to compute the coverage ratio reward introduced in Sec.~\ref{sec:method_opt}. The egocentric map $M_t$ and historical poses are encoded by LocoTransformer~\cite{yang2021learning}, and taken as the input of our policy network.

\subsection{Long-term Action Space}
\label{sec:method_action}
    Prior methods suffer from critical limitations in action space design for long-horizon exploration tasks. First, motion primitives (e.g., move forward 10cm)~\cite{chenlearning, chaplotlearning, ramakrishnan2020occupancy, yan2023anm} force reliance on a two-stage global-local pipeline, which introduces excessively costly simulation trials, severely degrading training efficiency. Furthermore, such fragmented trajectories inevitably compromise smoothness in real-world deployments. 
    Second, short-term action spaces~\cite{georgakis2022uncertainty,feng2024naruto} constrained to immediate neighborhoods demand simultaneous learning of collision avoidance and exploration under conservative behaviors, resulting in myopic policies that fail to reason about long-term environmental structures. 

    % \tianfan{It is still not clear what is our action space? Maybe add a figure to explain it?}
    To address these limitations, we propose an action space within the navigable area that allows distant but reachable long-term goals as atomic actions. The action space defines relative movement in the agent's local $SE(2)$ frame, parameterized as vectors ($\Delta x$, $\Delta y$, $\Delta \theta$). Compared to the local planner that outputs the motion primitives as action, a heuristic A* planner~\cite{hart1968formal} is used to plan a local trajectory to determine whether the goal is reachable. Note that we only capture observations at the long-term goals (i.e., keyframes) during training to avoid excess rounds of simulation. This design disentangles global exploration intent from local navigation, enabling agents to focus on high-level decision-making while offloading low-level path safety to an A*-based verifier. Specifically, for each predicted long-term goal, our framework employs the dynamically updating global map $G_t$ that indicates states among {occupied, free, unknown}, to verify its connectivity to the agent's current position via lightweight A* planning. Only goals with collision-free and navigable paths are considered safe, ensuring reachability without sacrificing exploration diversity. This innovation bridges the gap between reactive RL policies and deliberative planning, yielding trajectories that are both globally coherent and locally smooth.

\subsection{Training Strategy}
\label{sec:method_strategy}
    \noindent To enhance generalizability, we propose the following training strategies to diversify the decision process during policy learning, inspired by classic RL-based implementation~\cite{tobin2017domain, cobbe2020leveraging, li2022metadrive}. Also, we present additional termination conditions in our framework to accelerate policy learning. 

    \noindent\textbf{Scene Updating Strategy.}
    We leverage diverse training scenes from ProcTHOR dataset~\cite{deitke2022procthor} and AI2-THOR~\cite{kolve2022ai2thor} platform. However, we cannot launch such large numbers of parallel training environments in the simulator due to the limitations of computational efficiency and memory. On the other hand, Isaac Gym doesn't support users in replacing the loaded asset with a novel one. Thus, we cannot directly update the loaded scene using its provided API. In practice, we allocate $N//32$ distinct scenes per environment across 32 parallel Isaac Gym environments due to memory constraints, when training on $N (\geq512)$ scenes. During training, each environment may activate a scene from its inactive area with probability $p$ to replace the original one that would be moved to the inactive area. This design preserves scene diversity while maintaining memory efficiency.

    \noindent\textbf{Random Initialization Strategy.} Previous work~\cite{chaplotlearning} initializes the agent's position in the center of the scene, which implicitly introduces impractical prior to training, and thus cannot well generalize to unknown environments in the real world. To narrow this sim-to-real gap, we randomly set the initial poses of our agent in any non-collision area during training and evaluation. In particular, we extract the navigatable maps from preprocessed point clouds of scenes, and then slightly narrow the initial areas by max pooling to discard easy-to-collide corners. It turns out that the random initialization strategy significantly enriches the diversity of the decision process during training, and enhances the generalizability during evaluation.
    
    \noindent\textbf{Termination Conditions.} 
    Classic termination criteria include the maximum episode lengths and success thresholds. 
    To enhance exploration safety and navigation reliability in diverse environments, we propose three supplementary termination criteria: 1) \textit{Collision detection}, triggering immediate termination upon physical contact with obstacles; 2) \textit{Progress stagnation}, activated when the cumulative coverage reward over last ten steps falls below 1\%; and 3) \textit{Goal viability assessment}, terminating episodes where long-term objectives become unnavigable in the current observation space or exceed predefined trajectory length. These enhanced termination mechanisms collectively optimize risk mitigation while maintaining navigation effectiveness through adaptive response to the environment.

    \begin{table*}[!t]
    \small
    \center
    \vspace{-8pt}
    \setlength\tabcolsep{9pt}  % width of column
    \renewcommand{\arraystretch}{1.3}   % height of row
    \captionof{table}{The generalization results of exploration policies for active mapping on 128 unseen indoor scenes from our \oursbench, including synthetic datasets (\textbf{ProcTHOR}, \textbf{HSSD}) and real-scan datasets (\textbf{Gibson}, \textbf{Matterport3D}). All learnable methods are trained on 1024 scenes. We evaluate 10 episodes per scene and report the average results. Different methods initialize with the shared random poses of agents. ``*": ANM is based on per-scene optimized neural representation and thus is directly trained on each testing scene.}
    \begin{tabularx}{\textwidth}{ll|ccc|ccc|ccc}
        \hline
        % & \multirow{2}{*}{\textbf{Exploration Policy}} & \multicolumn{3}{c|}{\textbf{Overall}} & \multicolumn{3}{c|}{\textbf{ProcTHOR \& HSSD}} & \multicolumn{3}{c}{\textbf{Gibson \& Matterport3D}} \\ \cline{3-11}
        \multicolumn{2}{c|}{\multirow{2}{*}{\textbf{Exploration Policy}}} & \multicolumn{3}{c|}{\textbf{Overall}} & \multicolumn{3}{c|}{\textbf{ProcTHOR \& HSSD}} & \multicolumn{3}{c}{\textbf{Gibson \& Matterport3D}} \\ \cline{3-11}
        &  & \textbf{Cov.$~\uparrow$} & \textbf{AUC$~\uparrow$} & \textbf{CD~$\downarrow$} & \textbf{Cov.$~\uparrow$} & \textbf{AUC$~\uparrow$} & \textbf{CD~$\downarrow$} & \textbf{Cov.$~\uparrow$} & \textbf{AUC$~\uparrow$} & \textbf{CD~$\downarrow$} \\ \hline
    
        \parbox[c]{1.2mm}{\multirow{3}{*}{\rotatebox[origin=c]{90}{Heuristic~}}}
        & Random & 31.41\% & 27.58\% & 2.15m & 34.88\% & 30.85\% & 1.85m & 20.89\% & 24.30\% & 2.76m \\
        & Vacuum~\cite{chenlearning} & 42.96\% & 35.89\% & 1.67m & 45.81\% & 37.56\% & 1.41m & 37.13\% & 32.48\% & 2.20m \\ 
        & FBE~\cite{Yamauchi1997FBE} & 56.80\% & 45.07\% & 0.94m & 61.56\% & 50.99\% & 0.56m & 46.17\% & 37.60\% & 1.72m \\
        \hline
    
        \parbox[c]{1.2mm}{\multirow{2}{*}{\rotatebox[origin=c]{90}{Info~}}}
        & UPEN~\cite{georgakis2022uncertainty} & 49.65\% & 42.98\% & 1.38m & 54.84\% & 46.65\% & 1.01m & 39.02\% & 35.47\% & 2.14m \\
        & ANM*~\cite{yan2023anm} & 57.01\% & 49.56\% & 1.02m & 63.98\% & 56.37\% & 0.66m & 42.73\% & 35.63\% & 1.77m \\ 
        % & *NARUTO~\cite{feng2024naruto}  & \% & \% & m & \% & \% & m \\ 
        \hline
    
        \parbox[c]{1.2mm}{\multirow{3}{*}{\rotatebox[origin=c]{90}{RL-based~}}}
        & ANS~\cite{chaplotlearning} & 48.86\% & 41.98\% & 1.39m & 54.31\% & 46.93\% & 1.01m & 37.71\% & 31.86\% & 2.17m \\
        & OccAnt~\cite{ramakrishnan2020occupancy} & 53.61\% & 46.03\% & 1.16m & 60.30\% & 51.99\% & 0.82m & 39.91\% & 33.84\% & 1.85m \\
        & \cellcolor{gray!20}\textbf{\ours} & \cellcolor{gray!20}\textbf{66.50\%} & \cellcolor{gray!20}\textbf{57.63\%} & \cellcolor{gray!20}\textbf{0.80m} & \cellcolor{gray!20}\textbf{76.01\%} & \cellcolor{gray!20}\textbf{66.13\%} & \cellcolor{gray!20}\textbf{0.38m} & \cellcolor{gray!20}\textbf{47.04\%} & \cellcolor{gray!20}\textbf{40.23\%} & \cellcolor{gray!20}\textbf{1.67m} \\
        \hline
    \end{tabularx}
    \label{tab:main_table}
    % \vspace{-5pt}
    \end{table*}

\subsection{Reward Functions and Optimization}
\label{sec:method_opt}
    We highlight the optimization objective of our exploration policy lies in the following key aspects: exploration completeness, path efficiency, safety, and navigability.
    % ~\tianfan{Similarly, highlight what is special in our design. I assume that some algorithms only considers a subset of them?} 
    We design the following optimization settings, such as collision penalty, to encourage our exploration policy to predict reliable and safe target poses. The details of implementation can be found in Appendix~\ref{app:astar} and ~\ref{app:sceneupdate}.

    \noindent\textbf{The Setup of Reinforcement Learning.} Our end-to-end exploration policy is optimized with proximal policy optimization~\cite{schulman2017proximal} (PPO) for parallelizing sampling.
    Hence we design the following reward functions to reflect the task objective of exploration for active mapping.

    \noindent\textbf{Reward Functions.}
     With the occupancy probability $F^G_t$ at time step $t$, we can threshold each voxel with an empirical bound to determine if it is occupied.
     This discrimination process outputs a binary occupancy map with $\Tilde{N}_t$ voxels being occupied, which is used to calculate the coverage ratio: 
        \begin{equation}
        \mathrm{CR}_t = \frac{\Tilde{N}_t}{N^*} \cdot 100\% ,
        \end{equation}
    where $N^*$ is the number of ground-truth occupied cells representing the surface of scenes.
    % $N_{obs, t}$ is the number of observed occupied voxels among the ground-truth occupancy map at the time step $t$.
    To encourage our exploration policy to cover as many unseen scene areas as possible, we use the difference of coverage ratio (CR) between two consecutive steps as the main reward function $r^{CR}$:
    \begin{equation}
    r^\mathrm{CR}_{t+1} = \mathrm{CR}_{t+1} - \mathrm{CR}_t.
    \end{equation}

    To mitigate hazardous exploration behaviors, a negative reward is proposed to penalize collisions at long-term target viewpoints. When the predicted goal lies within the observed area and would cause a collision, the agent will remain stationary, re-plan its next goal, but still incur the collision penalty. Re-planning avoids unnecessary risky behavior and improves sample efficiency during training. In addition, the policy learning benefits from a termination reward triggered at 90\% coverage ratio, addressing the challenge of sparse successful samples in long-horizon exploration tasks. In particular, our collision reward $r_t^{\mathrm{Col}} = -1$ if a collision occurs at step t, else $0$. The termination reward $r_t^{\mathrm{Term}} = +1$ if the episode terminates with the final coverage exceeding $75\%$, else $0$.

    \begin{table}[!t]
    \center
    \setlength{\tabcolsep}{12pt}
    \renewcommand{\arraystretch}{1.3}
    \captionof{table}{Ablation studies of the diversity and complexity of training scenes. We train our exploration policies on different dataset sources to show the advances of cross-dataset and complex training data. The \textbf{bold} lines indicate the optimal designs, which are proposed by us and adopted in our framework.}
    \vspace{-7pt}
    \label{tab:ablation_data}
    \resizebox{1.0\linewidth}{!}{
    \begin{tabular}{cc|ccc}
    \toprule
        \textbf{\# of Scenes} & \textbf{Data Source} & \textbf{Cov.$~\uparrow$} & \textbf{AUC$~\uparrow$} & \textbf{CD$~\downarrow$}\\ \hline
        96 & Gibson & 50.32\% & 43.08\% & 1.09m \\
        192 & ProcTHOR ($\geq$ 10 rooms) & 63.23\% & 53.74\% & 0.90m \\
        416 & ProcTHOR ($<$ 6 rooms) & 61.66\% & 53.07\% & 0.91m \\
        896 & ProcTHOR & 65.50\% & 55.46\% & 0.83m  \\
        928 & ProcTHOR, HSSD & 66.09\% & 57.00\% & 0.81m\\
        \textbf{1024} & ProcTHOR, HSSD, Gibson & \textbf{66.50\%} & \textbf{57.63\%} & \textbf{0.80m} \\
        % \hline
        % \multicolumn{4}{c}{\textbf{Proportional Amount} \xiao{curve}} \\ \hline
        % 32 & 57.88\% & 48.12\% & 1.05m \\
        % 64 & 60.19\% & 51.34\% & 0.99m \\
        % 128 & 61.29\% & 52.33\% & 0.96m \\
        % 256 & 63.23\% & 54.06\% & 0.88m \\
        % 512 & 64.01\% & 54.77\% & 0.88m \\
        % \textbf{1024} & \textbf{\%} & \textbf{\%} & \textbf{0.m} \\
    \bottomrule
    \end{tabular}}
    \end{table}

    \begin{figure}[!t]
        \centering
        \vspace{-5pt}
        \includegraphics[width=0.48\textwidth]{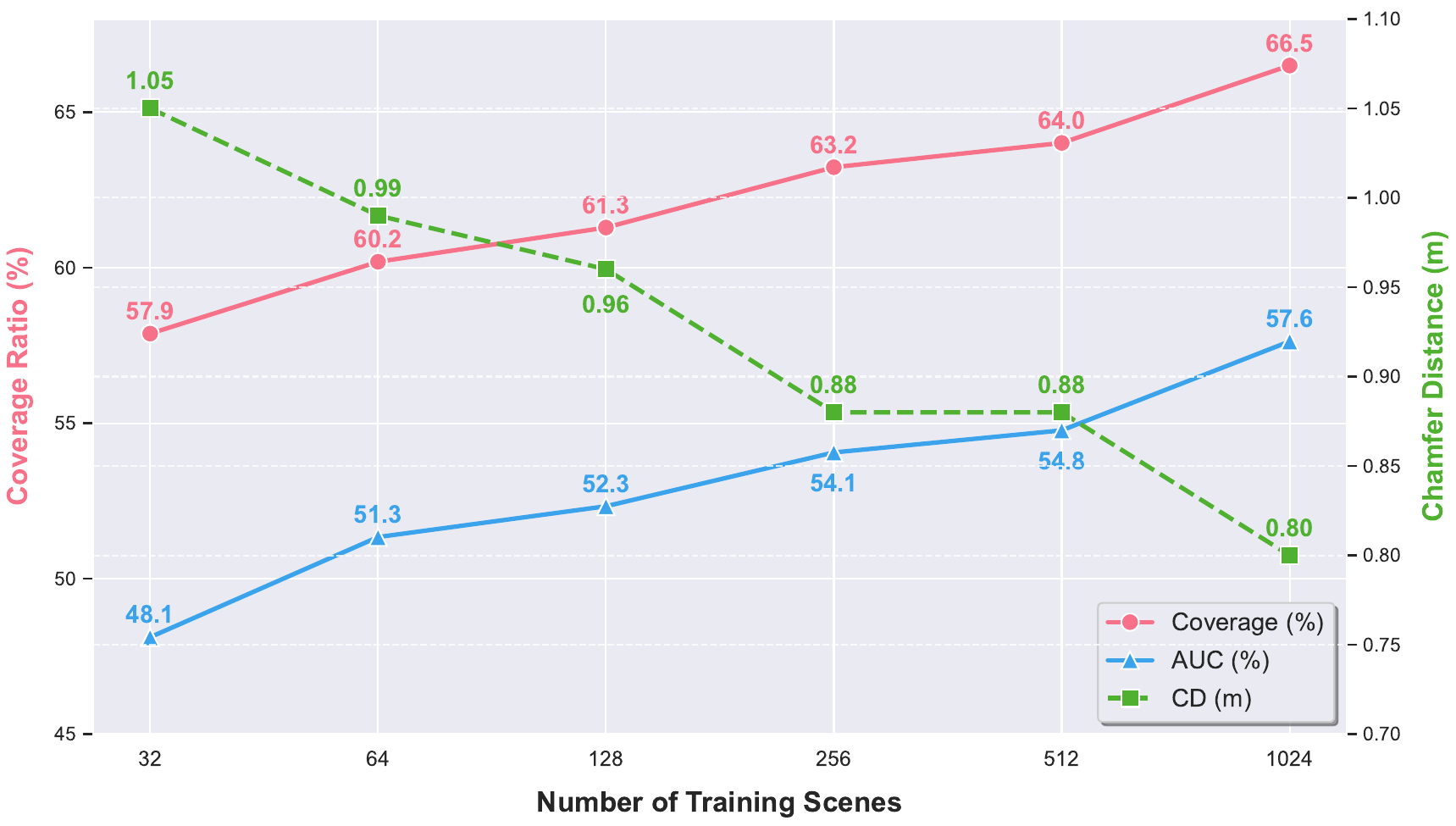}
        \caption{Ablation studies of the number of training scenes. We sample a proportionally reduced number of scenes from the complete 1024 training scenes to demonstrate the significance of the quantity and diversity of training scenes.}
        \label{fig:curve}   
        \vspace{-12pt}
    \end{figure}

\section{Experiments}
\label{sec:exp}

\subsection{Experimental Setup}
\label{sec:exp_setup}
    \noindent\textbf{Dataset.} 
    We conduct comprehensive training and evaluation of exploration policies on the proposed \oursbench benchmark. Specifically, the policy is trained using 1,024 indoor scenes spanning three datasets: two synthetic environments (ProcTHOR-10K~\cite{deitke2022procthor}, HSSD~\cite{khanna2024hssd}) and one real-scan dataset (Gibson~\cite{xia2018gibson}). To rigorously validate cross-domain generalizability, we generalize the unified policy to 128 unseen scenes from the mentioned three datasets and a challenging real-scan dataset, Matterport3D~\cite{chang2017matterport3d}, for zero-shot evaluation.
    The details of data sources, data preprocessing, and dataset split can be found in Appendix~\ref{app:data}.

    \noindent\textbf{Simulation Environment.}
    We conduct all experiments in NVIDIA Isaac Gym~\cite{makoviychuk2021isaac}, a GPU-accelerated physics simulation platform designed for embodied AI. 

    \noindent\textbf{Evaluation Metrics.} 
    We adopt the following metrics to evaluate the exploration completion, trajectory efficiency, and reconstruction accuracy for the active mapping task: (1) \textit{Coverage (\%)} quantifies the percentage of the environment successfully mapped by the end of the exploration process. It is calculated as the ratio of the number of explored points to the total number of ground-truth points. (2) \textit{AUC (\%)} measures the cumulative coverage ratio over the entire exploration duration, providing insight into both the path efficiency and thoroughness of exploration. (3) \textit{Chamfer Distance (CD, unit: meter)}~\cite{sucar2021imap,zhu2022nice} is the mono-directional Chamfer distance between each ground-truth point and the nearest captured points. 
    The upward arrow symbol ($\uparrow$) indicates that higher values correspond to superior performance in the evaluated metric, whereas the downward arrow ($\downarrow$) signifies scenarios where diminished magnitudes are preferable for optimal outcomes.

    \noindent\textbf{Implementation Details.} 
    All experiments are conducted based on legged gym~\cite{rudin2022learning} in NVIDIA Isaac Gym~\cite{makoviychuk2021isaac}, using CrazyFlie~\cite{giernacki2017crazyflie} as our agent equipped with an onboard depth camera. The depth maps are rendered at a resolution of $256 \times 256$ with a $90$\textdegree field of view (FOV). The resolution of our egocentric semantic map $M_t$ and probabilistic global map $G_t$ is both $128 \times 128$. The cell size of $M_t$ is $10cm$ $\times$ $10cm$. The evaluation is under the keyframe budget of $T=50$. The ground-truth point clouds on the surfaces are generated using the Poisson Disk sampling method~\cite{yuksel2015sample} through the Open3D API~\cite{Zhou2018}.
    The exploration policy is optimized through $5k$ iterations and uses approximately 96 hours of training time on a single GeForce RTX 4090 GPU. 
    All networks are randomly initialized. All networks are frozen during evaluation. 
    Please refer to the Appendix~\ref{app:details} for further results and details.

    \begin{table}[!t]
    \center
    \setlength{\tabcolsep}{12pt}
    \renewcommand{\arraystretch}{1.3}
    \captionof{table}{Ablation studies of the scene representations and training strategies. The \textbf{bold} lines indicate the optimal designs, which are proposed by us and adopted in our framework.}
    \vspace{-3pt}
    \label{tab:ablation_design}
    \resizebox{1.0\linewidth}{!}{
        \begin{tabular}{l|ccc}
        \toprule
            \textbf{\makecell{Settings}} & \textbf{Cov.$~\uparrow$} & \textbf{AUC$~\uparrow$} & \textbf{CD$~\downarrow$} \\
    
            \hline \multicolumn{4}{c}{\textbf{Scene Representation}} \\ \hline
             Binary Occupancy Map & 62.92\% & 53.98\% & 0.96m \\
             Probabilistic Occupancy Map & 64.37\% & 55.37\% & 0.83m \\
             \textbf{Semantic Occupancy Map} & \textbf{66.50\%} & \textbf{57.63\%} & \textbf{0.80m} \\
    
            \hline \multicolumn{4}{c}{\textbf{Feature Encoder}} \\ \hline
            UNet~\cite{ronneberger2015u,ramakrishnan2020occupancy} & 64.01\% & 54.77\% & 0.89m \\
            \textbf{LocoTransformer}~\cite{yang2021learning} & \textbf{66.50\%} & \textbf{57.63\%} & \textbf{0.80m} \\
    
            \hline \multicolumn{4}{c}{\textbf{Action Space}} \\ \hline
            Short-term Goal& 48.85\% & 40.24\% & 1.46m \\
            \textbf{Long-term Goal} & \textbf{66.50\%} & \textbf{57.63\%} & \textbf{0.80m} \\
    
            \hline \multicolumn{4}{c}{\textbf{Initialization Strategy}} \\ \hline
            Fixedly Initialize & 56.05\% & 47.95\% & 0.92m \\
            \textbf{Randomly Initialize} & \textbf{66.50\%} & \textbf{57.63\%} & \textbf{0.80m} \\
    
            \hline \multicolumn{4}{c}{\textbf{Scene Updating Strategy}} \\ \hline
            Updating Probability $p=0.05$ & 63.17\% & 53.90\% & 0.86m \\
            \textbf{Updating Probability $p=1$} & \textbf{66.50\%} & \textbf{57.63\%} & \textbf{0.80m} \\
        \bottomrule
        \end{tabular}
    }
    \vspace{-5pt}
    \end{table}
    
\subsection{Performance Comparison}
\label{sec:exp_result}
    To conduct a comprehensive evaluation of \ours's effectiveness and generalizability, we systematically assess its performance across both synthetic and real-world test scenarios, as detailed in Table~\ref{tab:main_table}. 
    % Our benchmark incorporates 128 evaluation samples spanning four distinct datasets, including a particularly challenging cross-dataset generalization test using real-world data from Matterport3D. 
    We implement and evaluate these works in our benchmark to demonstrate the superiority of our proposed method. The key details include: 
    1) \textbf{Random Policy} randomly samples actions from a Gaussian distribution within the action space. 2) \textbf{Vacuum Policy} simulates a heuristic exploration policy for robot vacuums. We follow~\cite{chenlearning} to let the policy move straight when safe and execute a random number of 9\degree turns when a collision occurs. 3) \textbf{FBE} always moves the agent towards the navigable nearest boundary between observed and unknown areas. 4) \textbf{UPEN}~\cite{georgakis2022uncertainty} estimates the information gain of candidate trajectories sampled by RRT~\cite{lavalle1998rapidly}, where the gain is estimated by model ensembles. We reduce the number of ensembling models and the number of layers due to the limited memory. 5) We replace \textbf{ANM}~\cite{yan2023anm}'s learned RL-based local planner with our A-star planner. 6) \textbf{ANS}~\cite{chaplotlearning} and \textbf{OccAnt}~\cite{ramakrishnan2020occupancy} are adapted to active mapping systems, given ground-truth poses. More details can be found in Appendix~\ref{app:details}.

    The experimental results in Table~\ref{tab:main_table} reveal \ours's superior generalizability across all evaluation metrics (Table~\ref{tab:main_table}). Our method achieves more than 9.49\% overall coverage improvement over existing baselines. In particular, our policy attains 76.01\% coverage in unseen synthetic indoor scenes that have more than five rooms on average. Notably, even for extremely challenging and cross-dataset real-scan scenes, our policy maintains around 50\% final coverage.

    To evaluate the deployability in real-world scenarios, we evaluate the exploration trajectory (unit: meter) and safety in Table~\ref{tab:keyframes}. It shows that \ours effectively explore unseen obstacle-dense scenarios while ensuring exploration safety.

    \begin{figure*}[!t]
        \hsize=\textwidth
        \centering
        % \vspace{-40pt}
        \includegraphics[width=\textwidth]{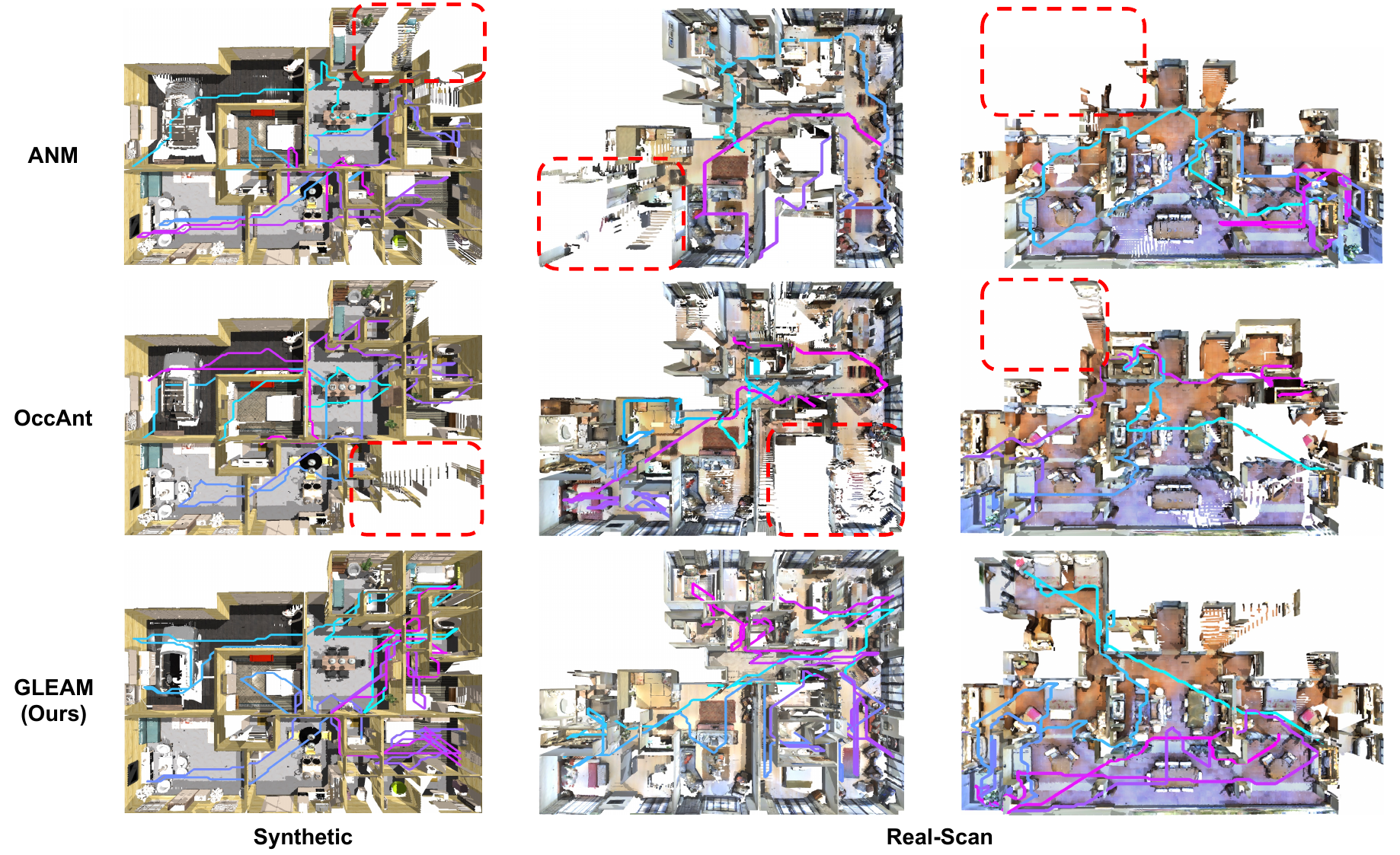}
        \caption{The visualization results of ANM, OccAnt, and \ours on three unseen complex indoor scenes from the test set of \oursbench. The methods share the same random initial poses for each scene.}
        \label{fig:vis}
        \vspace{-0.2cm}
    \end{figure*}

    \begin{table}[!h]
    \centering
    \footnotesize
    \vspace{-8pt}
    \setlength\tabcolsep{2.6pt}  % width of column
    \renewcommand{\arraystretch}{1.2}   % height of row
    \captionof{table}{The average number of keyframes during the generalization evaluation on 128 unseen indoor scenes from \oursbench.}
    \vspace{-7pt}
    \begin{tabularx}{\columnwidth}{l|cccccc}
        \hline
        % \textbf{Policy} & Random & FBE & UPEN & ANM & OccAnt & \ours \\ \hline
        \textbf{Policy} & \textbf{Random} & \textbf{FBE} & \textbf{UPEN} & \textbf{ANM} & \textbf{OccAnt} & \textbf{\ours} \\ \hline
        \textbf{Coverage} & 31.41\% & 56.80\% & 49.65\% & 57.01\% & 53.61\% & 66.50\% \\
        \textbf{\#Keyframes} & 37.01 & 26.21 & 24.84 & 28.31 & 18.58 & 29.57 \\
        \textbf{Traj. Length} & 9.56m & 52.34m & 30.45m & 46.51m & 38.08m & 54.51m \\
        \hline
    \end{tabularx}
    \label{tab:keyframes}
    \vspace{-9pt}
    \end{table}

\subsection{Ablation Study}
\label{sec:exp_ablation}
    \noindent\textbf{The Effect of Training Scenes.} As shown in Table~\ref{tab:ablation_data} and Fig.~\ref{fig:curve}, we demonstrate that training scene quantity, diversity, and complexity jointly enhance the generalization of exploration policy for active mapping. The policy trained on 96 Gibson scenes underperforms across all metrics, achieving only 50.32\% coverage.
    An inspiring result is that the policy trained on fewer but more complex scenes (with $\geq 10$ rooms) matches the performance of those using twice as many simple scenes, emphasizing the critical role of scene complexity. 
    Compared to the ProcTHOR-only baseline, our best policy integrating multi-domain datasets improves AUC and coverage by 2.17\% and 1.00\%, respectively, highlighting cross-dataset heterogeneity as critical for generalizability.

    Scaling training scenes from 32 to 1,024 monotonically improves coverage (+8.6\%), AUC (+9.5\%), and completion (-0.25m). This trend reveals that larger datasets force models to learn generalized exploration strategies by exposing them to rare spatial patterns (e.g., narrow corridors and multi-room connections). 
    % Trained on 1,024 scenes, combining multi-source data and scale achieves peak performance, proving synergistic effects between data quantity and diversity. 
    In summary, the shown results establish that sufficient, diverse, and complex training scenes are essential for robust active mapping in unseen environments, offering practical guidelines for learning generalizable robotic systems for active mapping.

    \noindent\textbf{The Effect of Scene Representations and Encoders.} We ablate the scene representations in Table~\ref{tab:ablation_design}. 
    ``Binary Occupancy Map" (occupied, others) and ``Probabilistic Occupancy Map" (occupied, free, unknown) denote the egocentric maps extracted from the corresponding intermediate representations introduced in Sec.~\ref{sec:method_map}.
    Our semantic occupancy map that extends binary occupancy with navigability and frontier outperforms classic binary occupancy maps by 3.65\% AUC and 3.58\% coverage. This demonstrates that incorporating semantic task-specific information (e.g., frontier categories) enables more informed exploration decisions compared to geometric-only representations. For feature encoding, the lightweight LocoTransformer encoder substantially surpasses conventional UNet architectures, leveraging cross-layer attention to capture long-range spatial dependencies.

    \noindent\textbf{The Effect of Long-Term Action Space.} 
    % Employing Long-term Goal planning outperforms short-term goal. It demonstrates our claim that long-term action space ensures path safety and reachability without sacrificing exploration diversity.
    Our experiments confirm that long-term goal planning achieves superior performance over short-horizon counterparts (see Table~\ref{tab:ablation_design}). This validates two critical advantages of our design: 1) The integration of A*-verified path connectivity inherently resolves the safety-reachability dilemma faced by traditional two-stage planners, with effective exploration (+17.39\% AUC); 2) Decoupling global exploration decisions from local obstacle avoidance preserves action space diversity, enabling the policy to discover non-myopic trajectories that cover more unknown regions (+17.65\%).
    Crucially, these gains do not sacrifice training efficiency.

    \noindent\textbf{The Effect of Training Strategies.} We ablate the key training strategies, including random initialization and scene updating, in Table~\ref{tab:ablation_design}. Strategic training configurations prove vital for balancing exploration efficiency and robustness.
    Notably, the random initialization strategy yields a 9.68\% AUC and 10.45\% coverage improvement over fixed initialization, highlighting that diverse starting conditions prevent overfitting to specific layouts. We define $p$ as the probability of replacing the original active training scenes with another training scene after each episode. The frequent update of the scene ($p=1$) surpasses the occasional update ($p=0.05$) with coverage of 3.33\% and AUC of 3.73\%. These results collectively establish that goal horizon, initialization diversity, and update frequency must be jointly optimized to achieve robust generalization.

\subsection{Qualitative Results}
\label{sec:exp_vis}
    We visualize the mapping results and the predicted scanning trajectories of ANM, OccAnt, and \ours within a single episode in Fig.~\ref{fig:vis}. It demonstrates that our \ours generalizes well, even in complex indoor scenes with $\geq$ 10 rooms and a large number of obstacles. More visualization results can be found on our project website.

    \vspace{5pt}
    \begin{figure}[!t]
        \centering
        \includegraphics[width=0.48\textwidth]{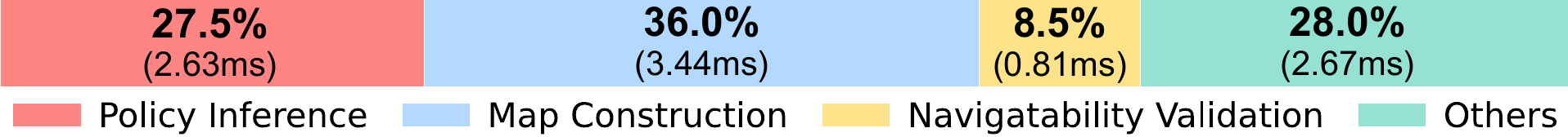}
        \setlength{\abovecaptionskip}{-6pt}
        \caption{The one-step inference time of our key components.}
        \vspace{-12pt}
        \label{fig:time}
    \end{figure}

    \begin{table}[!b]
    \center
    \footnotesize
    \setlength{\tabcolsep}{3pt}
    \renewcommand{\arraystretch}{1.05}
    \setlength{\belowcaptionskip}{-12pt}
    \captionof{table}{The robustness of \ours under Gaussian noise $N(0, \sigma^2)$ (unit: meter) during inference. \bm{$\sigma^2_P$}: variance of cumulative pose noise. \bm{$\sigma^2_D$}: variance of depth noise. \textbf{KF}: keyframes. \textbf{TL}: trajectory length. \textbf{``*"}: non-cumulative per-step noise}
    % \bm{$\mathcal{N_P}$}: cumulative pose noise. \bm{$\mathcal{N_D}$}: depth noise.
    \vspace{-8pt}
    \label{tab:noise_infer}
        \begin{tabular}{l|ll|ccccc}
        \hline
            \textbf{Settings} & \bm{$\sigma^2_P$} & \bm{$\sigma^2_D$} & \textbf{Cov.} & \textbf{AUC} & \textbf{CD} & \textbf{KF} & \textbf{TL} \\ \hline
            \rowcolor{gray!15}
            \multicolumn{3}{l|}{\ours (no noise)} & 66.50\% & 57.63\% & 0.80m & 29.57 & 54.51m \\ \hline
            \multirow{5}{*}{Pose-only} & $0.1$ & 0 & 63.27\% & 54.41\% & 0.86m & 28.27 & 41.63m \\
            & $0.3$ & 0 & 60.73\% & 52.11\% & 0.92m & 24.95 & 35.71m \\
            & $0.5$ & 0 & 55.59\% & 48.24\% & 0.99m & 20.31 & 22.79m \\
            & $0.1$* & 0 & 66.18\% & 56.62\% & 0.76m & 30.69 & 49.36m \\
            & $0.5$* & 0 & 58.54\% & 50.53\% & 0.95m & 23.61 & 38.51m \\
            \hline
            \multirow{3}{*}{Depth-only} & 0 & $0.05$ & 64.94\% & 55.31\% & 0.82m & 30.44 & 45.05m \\
            & 0 & $0.1$ & 60.21\% & 51.09\% & 0.96m & 29.21 & 36.78m \\
            & 0 & $0.2$ & 54.77\% & 46.30\% & 1.13m & 27.51 & 27.21m \\
            \hline
            \multirow{3}{*}{Depth+Pose} & $0.1$ & $0.05$ & 60.44\% & 51.76\% & 0.91m & 28.20 & 36.62m \\
            & $0.3$ & $0.1$ & 54.03\% & 47.21\% & 1.09m & 24.74 & 24.33m \\
            & $0.5$ & $0.1$ & 51.36\% & 44.32\% & 1.17m & 20.94 & 20.56m \\
            \hline
        \end{tabular}
    % \vspace{-10pt}
    \end{table}

\subsection{Discussion and Future Directions}
\label{sec:exp_dis}
    \noindent\textbf{Realistic settings \& Real-world deployment.} 1) \textit{Noisy observation.} Real-world deployments inevitably confront imperfect sensor inputs such as camera noise and depth ambiguity. Our framework employs probabilistic occupancy maps to mitigate noisy raw inputs of sensors, yet persistent noise patterns still propagate geometric errors during active mapping. 2) \textit{Pose estimation.} While our system obtains accurate poses through the simulator, the pose estimation in practical scenarios with fast camera motions or textureless regions induces pose drift. This spatial uncertainty manifests as misaligned geometry fragments, particularly when scanning thin structures like chair legs or lamp arms. 3) \textit{Open environments.} Unlike bounded scanning domains in simulation, real-world scenes often contain dynamically expanding areas (e.g., newly opened doors). Existing frameworks struggle to build memory-efficient representations for unbounded and dynamic scenes, which shows that the sim-to-real gap still has a lot of potential to be explored.

    While real-world deployment remains an open challenge, we advance the sim-to-real validation across \textit{sensor-noise tolerance} and \textit{computation cost} to demonstrate concrete progress toward deployability. To evaluate the sensor-noise tolerance of \ours, we inject hardware-aligned Gaussian noise to observations during inference, deriving from real-world sensors like Intel\textregistered RealSense D455 depth camera ($<2\%$ error at $4m$) and TDK InvenSense ICM-42688-P IMU. As shown in Table~\ref{tab:noise_infer}, \ours maintains strong robustness despite training on ideal observations, which stems from our probabilistic map that inherently suppresses transient noise by Bayesian updating. 

    Our system achieves real-time inference (\textbf{104.7Hz}) on a PC with an RTX 3090 GPU, with latency analysis in Figure~\ref{fig:time} demonstrating the efficiency of our lightweight policy network and CUDA-accelerated map updating/A* planning, ensuring seamless high-frequency perception and decision-making in the real world.

    \noindent\textbf{Challenging 3D benchmark \& 3D action space.} We've explored the potential of our framework for challenging 3D benchmark, including 3D action space ($x, y, z, pitch, yaw$) and 3D optimization objectives. In this setting, our agent is encouraged to capture all details, such as the surface under the underside of a table, in complex 3D scenes. While our agent demonstrates promising effectiveness and generalizability in scanning most of unobstructed surfaces, we found it quite difficult to capture the geometrically complex surfaces (e.g., the undersides of tables and self-occlusion surfaces of decorations) in cluttered 3D environments. Also, the heavy computational burden and limited memory prevent us from optimizing the components like 3D representations and scaling the number of training scenes.    
    
    \noindent\textbf{Challenging multi-floor complex scenes.} Although we've been exploring active mapping for complex single-floor indoor scenes, the tough cases in the real world are the multi-floor indoor scenes. These environments introduce unique cross-floor topological dependencies and vertical navigation constraints that existing frameworks fail to adequately model. Moreover, the inherent geometric discontinuities between floors exacerbate memory fragmentation when using conventional spatial representations, leading to increasing memory overhead.
    % To promote community development, we will also release 112 multi-floor high-quality scene meshes, manually selected from real-scan Gibson and Matterport3D and preprocessed by us.
    % 94 Gibson + 18 MP3D

    \noindent\textbf{Multi-agent collaboration.} Multi-agent collaboration is one of the solutions for complex scenarios such as multi-floor scenes. However, scaling to collaborative active mapping introduces novel fundamental challenges in distributed strategy optimization, dynamic role allocation, and communications. These challenges demand the rethinking of existing frameworks, particularly in developing scalable and memory-efficient representations and decentralized decision-making architectures.

\section{Conclusion}
    This work presents \oursbench, the first large-scale benchmark with 1,152 diverse 3D scenes to enable scalable training and reliable evaluation. 
    Furthermore, we propose \ours, an RL-based generalizable exploration policy for active mapping in complex unknown environments. It significantly outperforms previous methods, achieving 66.50\% coverage (+9.49\%) with efficient trajectories, and improved mapping accuracy on 128 unseen complex scenes. We show the effectiveness of our proposed modules, including semantic map representations and the randomized initialization strategy.

\section*{Acknowledgements}
    This work is supported in part by the Centre for Perceptual and Interactive Intelligence (CPII) Ltd., a CUHK-led InnoCentre under the InnoHK initiative of the Innovation and Technology Commission of the Hong Kong Special Administrative Region Government, the National Key R\&D Program of China (2022ZD0160201), and Shanghai Artificial Intelligence Laboratory.
    We thank all anonymous reviewers for their constructive feedback and valuable suggestions.
    We also extend our sincere gratitude to Yaokun Li, Chenjian Gao, and Ruikang Li for their advice on visualization, as well as to Gang Chen for his insightful guidance regarding drone control.

\clearpage
{
    \small
    \bibliographystyle{ieeenat_fullname}
    \bibliography{main}

\begin{thebibliography}{55}
\providecommand{\natexlab}[1]{#1}
\providecommand{\url}[1]{\texttt{#1}}
\expandafter\ifx\csname urlstyle\endcsname\relax
  \providecommand{\doi}[1]{doi: #1}\else
  \providecommand{\doi}{doi: \begingroup \urlstyle{rm}\Url}\fi

\bibitem[lav(1998)]{lavalle1998rapidly}
Rapidly-exploring random trees: A new tool for path planning.
\newblock \emph{Research Report 9811}, 1998.

\bibitem[Batinovic et~al.(2021)Batinovic, Petrovic, Ivanovic, Petric, and Bogdan]{batinovic2021multi}
Ana Batinovic, Tamara Petrovic, Antun Ivanovic, Frano Petric, and Stjepan Bogdan.
\newblock A multi-resolution frontier-based planner for autonomous 3d exploration.
\newblock \emph{IEEE Robotics and Automation Letters}, 6\penalty0 (3):\penalty0 4528--4535, 2021.

\bibitem[Bircher et~al.(2016)Bircher, Kamel, Alexis, Oleynikova, and Siegwart]{bircher2016receding}
Andreas Bircher, Mina Kamel, Kostas Alexis, Helen Oleynikova, and Roland Siegwart.
\newblock Receding horizon" next-best-view" planner for 3d exploration.
\newblock In \emph{2016 IEEE international conference on robotics and automation (ICRA)}, pages 1462--1468. IEEE, 2016.

\bibitem[Bresenham(1965)]{bresenham1965algorithm}
Jack~E Bresenham.
\newblock Algorithm for computer control of a digital plotter.
\newblock \emph{IBM Systems journal}, 4\penalty0 (1):\penalty0 25--30, 1965.

\bibitem[Cao et~al.(2024)Cao, Zhao, Wang, Xiang, and Sartoretti]{cao2024deep}
Yuhong Cao, Rui Zhao, Yizhuo Wang, Bairan Xiang, and Guillaume Sartoretti.
\newblock Deep reinforcement learning-based large-scale robot exploration.
\newblock \emph{IEEE Robotics and Automation Letters}, 2024.

\bibitem[Chang et~al.(2017)Chang, Dai, Funkhouser, Halber, Niessner, Savva, Song, Zeng, and Zhang]{chang2017matterport3d}
Angel Chang, Angela Dai, Thomas Funkhouser, Maciej Halber, Matthias Niessner, Manolis Savva, Shuran Song, Andy Zeng, and Yinda Zhang.
\newblock Matterport3d: Learning from rgb-d data in indoor environments.
\newblock \emph{arXiv preprint arXiv:1709.06158}, 2017.

\bibitem[Chaplot et~al.(2020)Chaplot, Gandhi, Gupta, Gupta, and Salakhutdinov]{chaplotlearning}
Devendra~Singh Chaplot, Dhiraj Gandhi, Saurabh Gupta, Abhinav Gupta, and Ruslan Salakhutdinov.
\newblock Learning to explore using active neural slam.
\newblock In \emph{International Conference on Learning Representations}, 2020.

\bibitem[Chen et~al.(2019)Chen, Gupta, and Gupta]{chenlearning}
Tao Chen, Saurabh Gupta, and Abhinav Gupta.
\newblock Learning exploration policies for navigation.
\newblock In \emph{International Conference on Learning Representations}, 2019.

\bibitem[Chen et~al.(2024)Chen, Li, Wang, Xue, and Pang]{chen2024gennbv}
Xiao Chen, Quanyi Li, Tai Wang, Tianfan Xue, and Jiangmiao Pang.
\newblock Gennbv: Generalizable next-best-view policy for active 3d reconstruction.
\newblock In \emph{Proceedings of the IEEE/CVF Conference on Computer Vision and Pattern Recognition}, pages 16436--16445, 2024.

\bibitem[Cobbe et~al.(2020)Cobbe, Hesse, Hilton, and Schulman]{cobbe2020leveraging}
Karl Cobbe, Chris Hesse, Jacob Hilton, and John Schulman.
\newblock Leveraging procedural generation to benchmark reinforcement learning.
\newblock In \emph{International conference on machine learning}, pages 2048--2056. PMLR, 2020.

\bibitem[Dai et~al.(2017)Dai, Chang, Savva, Halber, Funkhouser, and Nie{\ss}ner]{dai2017scannet}
Angela Dai, Angel~X Chang, Manolis Savva, Maciej Halber, Thomas Funkhouser, and Matthias Nie{\ss}ner.
\newblock Scannet: Richly-annotated 3d reconstructions of indoor scenes.
\newblock In \emph{Proceedings of the IEEE conference on computer vision and pattern recognition}, pages 5828--5839, 2017.

\bibitem[Deitke et~al.(2022)Deitke, VanderBilt, Herrasti, Weihs, Ehsani, Salvador, Han, Kolve, Kembhavi, and Mottaghi]{deitke2022procthor}
Matt Deitke, Eli VanderBilt, Alvaro Herrasti, Luca Weihs, Kiana Ehsani, Jordi Salvador, Winson Han, Eric Kolve, Aniruddha Kembhavi, and Roozbeh Mottaghi.
\newblock Procthor: Large-scale embodied ai using procedural generation.
\newblock \emph{Advances in Neural Information Processing Systems}, 35:\penalty0 5982--5994, 2022.

\bibitem[Dornhege and Kleiner(2013)]{dornhege2013frontier}
Christian Dornhege and Alexander Kleiner.
\newblock A frontier-void-based approach for autonomous exploration in 3d.
\newblock \emph{Advanced Robotics}, 27\penalty0 (6):\penalty0 459--468, 2013.

\bibitem[Feng et~al.(2024)Feng, Zhan, Chen, Yan, Xu, Cai, Li, Zhu, and Xu]{feng2024naruto}
Ziyue Feng, Huangying Zhan, Zheng Chen, Qingan Yan, Xiangyu Xu, Changjiang Cai, Bing Li, Qilun Zhu, and Yi Xu.
\newblock Naruto: Neural active reconstruction from uncertain target observations.
\newblock In \emph{Proceedings of the IEEE/CVF Conference on Computer Vision and Pattern Recognition}, pages 21572--21583, 2024.

\bibitem[Georgakis et~al.(2022)Georgakis, Bucher, Arapin, Schmeckpeper, Matni, and Daniilidis]{georgakis2022uncertainty}
Georgios Georgakis, Bernadette Bucher, Anton Arapin, Karl Schmeckpeper, Nikolai Matni, and Kostas Daniilidis.
\newblock Uncertainty-driven planner for exploration and navigation.
\newblock In \emph{2022 International Conference on Robotics and Automation (ICRA)}, pages 11295--11302. IEEE, 2022.

\bibitem[Giernacki et~al.(2017)Giernacki, Skwierczy{\'n}ski, Witwicki, Wro{\'n}ski, and Kozierski]{giernacki2017crazyflie}
Wojciech Giernacki, Mateusz Skwierczy{\'n}ski, Wojciech Witwicki, Pawe{\l} Wro{\'n}ski, and Piotr Kozierski.
\newblock Crazyflie 2.0 quadrotor as a platform for research and education in robotics and control engineering.
\newblock In \emph{2017 22nd International Conference on Methods and Models in Automation and Robotics (MMAR)}, pages 37--42. IEEE, 2017.

\bibitem[Hart et~al.(1968)Hart, Nilsson, and Raphael]{hart1968formal}
Peter~E Hart, Nils~J Nilsson, and Bertram Raphael.
\newblock A formal basis for the heuristic determination of minimum cost paths.
\newblock \emph{IEEE transactions on Systems Science and Cybernetics}, 4\penalty0 (2):\penalty0 100--107, 1968.

\bibitem[Isler et~al.(2016)Isler, Sabzevari, Delmerico, and Scaramuzza]{isler2016information}
Stefan Isler, Reza Sabzevari, Jeffrey Delmerico, and Davide Scaramuzza.
\newblock An information gain formulation for active volumetric 3d reconstruction.
\newblock In \emph{2016 IEEE International Conference on Robotics and Automation (ICRA)}, pages 3477--3484. IEEE, 2016.

\bibitem[Jiang et~al.(2023)Jiang, Lei, and Daniilidis]{jiang2023fisherrf}
Wen Jiang, Boshu Lei, and Kostas Daniilidis.
\newblock Fisherrf: Active view selection and uncertainty quantification for radiance fields using fisher information.
\newblock \emph{arXiv preprint arXiv:2311.17874}, 2023.

\bibitem[Juliani et~al.(2018)Juliani, Berges, Vckay, Gao, Henry, Mattar, and Lange]{juliani2020unity}
Arthur Juliani, Vincent-Pierre Berges, Esh Vckay, Yuan Gao, Hunter Henry, Marwan Mattar, and Danny Lange.
\newblock Unity: A general platform for intelligent agents.
\newblock \emph{arXiv preprint arXiv:1809.02627}, 2018.

\bibitem[Kerbl et~al.(2023)Kerbl, Kopanas, Leimk{\"u}hler, and Drettakis]{kerbl20233d}
Bernhard Kerbl, Georgios Kopanas, Thomas Leimk{\"u}hler, and George Drettakis.
\newblock 3d gaussian splatting for real-time radiance field rendering.
\newblock \emph{ACM Trans. Graph.}, 42\penalty0 (4):\penalty0 139--1, 2023.

\bibitem[Khanna et~al.(2024)Khanna, Mao, Jiang, Haresh, Shacklett, Batra, Clegg, Undersander, Chang, and Savva]{khanna2024hssd}
Mukul Khanna, Yongsen Mao, Hanxiao Jiang, Sanjay Haresh, Brennan Shacklett, Dhruv Batra, Alexander Clegg, Eric Undersander, Angel~X Chang, and Manolis Savva.
\newblock Habitat synthetic scenes dataset (hssd-200): An analysis of 3d scene scale and realism tradeoffs for objectgoal navigation.
\newblock In \emph{Proceedings of the IEEE/CVF Conference on Computer Vision and Pattern Recognition}, pages 16384--16393, 2024.

\bibitem[Kl{\"o}ckner et~al.(2012)Kl{\"o}ckner, Pinto, Lee, Catanzaro, Ivanov, and Fasih]{klockner2012pycuda}
Andreas Kl{\"o}ckner, Nicolas Pinto, Yunsup Lee, Bryan Catanzaro, Paul Ivanov, and Ahmed Fasih.
\newblock Pycuda and pyopencl: A scripting-based approach to gpu run-time code generation.
\newblock \emph{Parallel Computing}, 38\penalty0 (3):\penalty0 157--174, 2012.

\bibitem[Kolve et~al.(2017)Kolve, Mottaghi, Han, VanderBilt, Weihs, Herrasti, Deitke, Ehsani, Gordon, Zhu, et~al.]{kolve2022ai2thor}
Eric Kolve, Roozbeh Mottaghi, Winson Han, Eli VanderBilt, Luca Weihs, Alvaro Herrasti, Matt Deitke, Kiana Ehsani, Daniel Gordon, Yuke Zhu, et~al.
\newblock Ai2-thor: An interactive 3d environment for visual ai.
\newblock \emph{arXiv preprint arXiv:1712.05474}, 2017.

\bibitem[Lee et~al.(2022)Lee, Chen, Wang, Liniger, Kumar, and Yu]{lee2022uncertainty}
Soomin Lee, Le Chen, Jiahao Wang, Alexander Liniger, Suryansh Kumar, and Fisher Yu.
\newblock Uncertainty guided policy for active robotic 3d reconstruction using neural radiance fields.
\newblock \emph{IEEE Robotics and Automation Letters}, 7\penalty0 (4):\penalty0 12070--12077, 2022.

\bibitem[Li et~al.(2022)Li, Peng, Feng, Zhang, Xue, and Zhou]{li2022metadrive}
Quanyi Li, Zhenghao Peng, Lan Feng, Qihang Zhang, Zhenghai Xue, and Bolei Zhou.
\newblock Metadrive: Composing diverse driving scenarios for generalizable reinforcement learning.
\newblock \emph{IEEE transactions on pattern analysis and machine intelligence}, 45\penalty0 (3):\penalty0 3461--3475, 2022.

\bibitem[Li et~al.(2025)Li, Gu{\'e}don, Boittiaux, Chen, and Lepetit]{li2025nextbestpath}
Shiyao Li, Antoine Gu{\'e}don, Cl{\'e}mentin Boittiaux, Shizhe Chen, and Vincent Lepetit.
\newblock Nextbestpath: Efficient 3d mapping of unseen environments.
\newblock \emph{arXiv preprint arXiv:2502.05378}, 2025.

\bibitem[Lluvia et~al.(2021)Lluvia, Lazkano, and Ansuategi]{lluvia2021active}
Iker Lluvia, Elena Lazkano, and Ander Ansuategi.
\newblock Active mapping and robot exploration: A survey.
\newblock \emph{Sensors}, 21\penalty0 (7):\penalty0 2445, 2021.

\bibitem[Makoviychuk et~al.(2021)Makoviychuk, Wawrzyniak, Guo, Lu, Storey, Macklin, Hoeller, Rudin, Allshire, Handa, et~al.]{makoviychuk2021isaac}
Viktor Makoviychuk, Lukasz Wawrzyniak, Yunrong Guo, Michelle Lu, Kier Storey, Miles Macklin, David Hoeller, Nikita Rudin, Arthur Allshire, Ankur Handa, et~al.
\newblock Isaac gym: High performance gpu-based physics simulation for robot learning.
\newblock \emph{arXiv preprint arXiv:2108.10470}, 2021.

\bibitem[Mildenhall et~al.(2020)Mildenhall, Srinivasan, Tancik, Barron, Ramamoorthi, and Ng]{mildenhall2020nerf}
Ben Mildenhall, Pratul~P Srinivasan, Matthew Tancik, Jonathan~T Barron, Ravi Ramamoorthi, and Ren Ng.
\newblock Nerf: Representing scenes as neural radiance fields for view synthesis.
\newblock In \emph{Computer Vision--ECCV 2020: 16th European Conference, Glasgow, UK, August 23--28, 2020, Proceedings, Part I 16}, pages 405--421. Springer, 2020.

\bibitem[Ortiz et~al.(2022)Ortiz, Clegg, Dong, Sucar, Novotny, Zollhoefer, and Mukadam]{iSDF2022}
Joseph Ortiz, Alexander Clegg, Jing Dong, Edgar Sucar, David Novotny, Michael Zollhoefer, and Mustafa Mukadam.
\newblock isdf: Real-time neural signed distance fields for robot perception.
\newblock \emph{Robotics: Science and Systems}, 2022.

\bibitem[Park et~al.(2019)Park, Florence, Straub, Newcombe, and Lovegrove]{park2019deepsdf}
Jeong~Joon Park, Peter Florence, Julian Straub, Richard Newcombe, and Steven Lovegrove.
\newblock Deepsdf: Learning continuous signed distance functions for shape representation.
\newblock In \emph{Proceedings of the IEEE/CVF conference on computer vision and pattern recognition}, pages 165--174, 2019.

\bibitem[Paszke et~al.(2019)Paszke, Gross, Massa, Lerer, Bradbury, Chanan, Killeen, Lin, Gimelshein, Antiga, et~al.]{paszke2019pytorch}
Adam Paszke, Sam Gross, Francisco Massa, Adam Lerer, James Bradbury, Gregory Chanan, Trevor Killeen, Zeming Lin, Natalia Gimelshein, Luca Antiga, et~al.
\newblock Pytorch: An imperative style, high-performance deep learning library.
\newblock \emph{Advances in neural information processing systems}, 32, 2019.

\bibitem[Raffin et~al.(2021)Raffin, Hill, Gleave, Kanervisto, Ernestus, and Dormann]{raffin2021stable}
Antonin Raffin, Ashley Hill, Adam Gleave, Anssi Kanervisto, Maximilian Ernestus, and Noah Dormann.
\newblock Stable-baselines3: Reliable reinforcement learning implementations.
\newblock \emph{The Journal of Machine Learning Research}, 22\penalty0 (1):\penalty0 12348--12355, 2021.

\bibitem[Ramakrishnan et~al.(2020)Ramakrishnan, Al-Halah, and Grauman]{ramakrishnan2020occupancy}
Santhosh~K Ramakrishnan, Ziad Al-Halah, and Kristen Grauman.
\newblock Occupancy anticipation for efficient exploration and navigation.
\newblock In \emph{Computer Vision--ECCV 2020: 16th European Conference, Glasgow, UK, August 23--28, 2020, Proceedings, Part V 16}, pages 400--418. Springer, 2020.

\bibitem[Ran et~al.(2023)Ran, Zeng, He, Chen, Li, Chen, Lee, and Ye]{ran2023neurar}
Yunlong Ran, Jing Zeng, Shibo He, Jiming Chen, Lincheng Li, Yingfeng Chen, Gimhee Lee, and Qi Ye.
\newblock Neurar: Neural uncertainty for autonomous 3d reconstruction with implicit neural representations.
\newblock \emph{IEEE Robotics and Automation Letters}, 2023.

\bibitem[Ronneberger et~al.(2015)Ronneberger, Fischer, and Brox]{ronneberger2015u}
Olaf Ronneberger, Philipp Fischer, and Thomas Brox.
\newblock U-net: Convolutional networks for biomedical image segmentation.
\newblock In \emph{Medical image computing and computer-assisted intervention--MICCAI 2015: 18th international conference, Munich, Germany, October 5-9, 2015, proceedings, part III 18}, pages 234--241. Springer, 2015.

\bibitem[Rudin et~al.(2021)Rudin, Hoeller, Reist, and Hutter]{rudinlearning}
Nikita Rudin, David Hoeller, Philipp Reist, and Marco Hutter.
\newblock Learning to walk in minutes using massively parallel deep reinforcement learning.
\newblock In \emph{5th Annual Conference on Robot Learning}, 2021.

\bibitem[Savva et~al.(2019)Savva, Kadian, Maksymets, Zhao, Wijmans, Jain, Straub, Liu, Koltun, Malik, et~al.]{savva2019habitat}
Manolis Savva, Abhishek Kadian, Oleksandr Maksymets, Yili Zhao, Erik Wijmans, Bhavana Jain, Julian Straub, Jia Liu, Vladlen Koltun, Jitendra Malik, et~al.
\newblock Habitat: A platform for embodied ai research.
\newblock In \emph{Proceedings of the IEEE/CVF international conference on computer vision}, pages 9339--9347, 2019.

\bibitem[Schulman et~al.(2017)Schulman, Wolski, Dhariwal, Radford, and Klimov]{schulman2017proximal}
John Schulman, Filip Wolski, Prafulla Dhariwal, Alec Radford, and Oleg Klimov.
\newblock Proximal policy optimization algorithms.
\newblock \emph{arXiv preprint arXiv:1707.06347}, 2017.

\bibitem[Song et~al.(2017)Song, Yu, Zeng, Chang, Savva, and Funkhouser]{song2017semantic}
Shuran Song, Fisher Yu, Andy Zeng, Angel~X Chang, Manolis Savva, and Thomas Funkhouser.
\newblock Semantic scene completion from a single depth image.
\newblock In \emph{Proceedings of the IEEE conference on computer vision and pattern recognition}, pages 1746--1754, 2017.

\bibitem[Straub et~al.(2019)Straub, Whelan, Ma, Chen, Wijmans, Green, Engel, Mur-Artal, Ren, Verma, et~al.]{straub2019replica}
Julian Straub, Thomas Whelan, Lingni Ma, Yufan Chen, Erik Wijmans, Simon Green, Jakob~J Engel, Raul Mur-Artal, Carl Ren, Shobhit Verma, et~al.
\newblock The replica dataset: A digital replica of indoor spaces.
\newblock \emph{arXiv preprint arXiv:1906.05797}, 2019.

\bibitem[Sucar et~al.(2021)Sucar, Liu, Ortiz, and Davison]{sucar2021imap}
Edgar Sucar, Shikun Liu, Joseph Ortiz, and Andrew~J Davison.
\newblock imap: Implicit mapping and positioning in real-time.
\newblock In \emph{Proceedings of the IEEE/CVF International Conference on Computer Vision}, pages 6229--6238, 2021.

\bibitem[Thrun(2002)]{thrun2002probabilistic}
Sebastian Thrun.
\newblock Probabilistic robotics.
\newblock \emph{Communications of the ACM}, 45\penalty0 (3):\penalty0 52--57, 2002.

\bibitem[Tobin et~al.(2017)Tobin, Fong, Ray, Schneider, Zaremba, and Abbeel]{tobin2017domain}
Josh Tobin, Rachel Fong, Alex Ray, Jonas Schneider, Wojciech Zaremba, and Pieter Abbeel.
\newblock Domain randomization for transferring deep neural networks from simulation to the real world.
\newblock In \emph{2017 IEEE/RSJ international conference on intelligent robots and systems (IROS)}, pages 23--30. IEEE, 2017.

\bibitem[Wang et~al.(2024)Wang, Mao, Zhu, Xu, Lyu, Li, Chen, Zhang, Chen, Xue, et~al.]{wang2024embodiedscan}
Tai Wang, Xiaohan Mao, Chenming Zhu, Runsen Xu, Ruiyuan Lyu, Peisen Li, Xiao Chen, Wenwei Zhang, Kai Chen, Tianfan Xue, et~al.
\newblock Embodiedscan: A holistic multi-modal 3d perception suite towards embodied ai.
\newblock In \emph{Proceedings of the IEEE/CVF Conference on Computer Vision and Pattern Recognition}, pages 19757--19767, 2024.

\bibitem[Xia et~al.(2018)Xia, Zamir, He, Sax, Malik, and Savarese]{xia2018gibson}
Fei Xia, Amir~R Zamir, Zhiyang He, Alexander Sax, Jitendra Malik, and Silvio Savarese.
\newblock Gibson env: Real-world perception for embodied agents.
\newblock In \emph{Proceedings of the IEEE conference on computer vision and pattern recognition}, pages 9068--9079, 2018.

\bibitem[Yamauchi(1997)]{Yamauchi1997FBE}
B. Yamauchi.
\newblock A frontier-based approach for autonomous exploration.
\newblock In \emph{Proceedings 1997 IEEE International Symposium on Computational Intelligence in Robotics and Automation CIRA'97. 'Towards New Computational Principles for Robotics and Automation'}, pages 146--151, 1997.

\bibitem[Yan et~al.(2023)Yan, Yang, and Zha]{yan2023anm}
Zike Yan, Haoxiang Yang, and Hongbin Zha.
\newblock Active neural mapping.
\newblock In \emph{Proceedings of the IEEE/CVF International Conference on Computer Vision}, pages 10981--10992, 2023.

\bibitem[Yang et~al.(2021)Yang, Zhang, Hansen, Xu, and Wang]{yang2021learning}
Ruihan Yang, Minghao Zhang, Nicklas Hansen, Huazhe Xu, and Xiaolong Wang.
\newblock Learning vision-guided quadrupedal locomotion end-to-end with cross-modal transformers.
\newblock \emph{arXiv preprint arXiv:2107.03996}, 2021.

\bibitem[Yeshwanth et~al.(2023)Yeshwanth, Liu, Nie{\ss}ner, and Dai]{yeshwanth2023scannet++}
Chandan Yeshwanth, Yueh-Cheng Liu, Matthias Nie{\ss}ner, and Angela Dai.
\newblock Scannet++: A high-fidelity dataset of 3d indoor scenes.
\newblock In \emph{Proceedings of the IEEE/CVF International Conference on Computer Vision}, pages 12--22, 2023.

\bibitem[Yuksel(2015)]{yuksel2015sample}
Cem Yuksel.
\newblock Sample elimination for generating poisson disk sample sets.
\newblock \emph{Computer Graphics Forum}, 34\penalty0 (2):\penalty0 25--32, 2015.

\bibitem[Zhan et~al.(2022)Zhan, Zheng, Xu, Reid, and Rezatofighi]{zhan2022activermap}
Huangying Zhan, Jiyang Zheng, Yi Xu, Ian Reid, and Hamid Rezatofighi.
\newblock Activermap: Radiance field for active mapping and planning.
\newblock \emph{arXiv preprint arXiv:2211.12656}, 2022.

\bibitem[Zhou et~al.(2018)Zhou, Park, and Koltun]{Zhou2018}
Qian-Yi Zhou, Jaesik Park, and Vladlen Koltun.
\newblock {Open3D}: {A} modern library for {3D} data processing.
\newblock \emph{arXiv:1801.09847}, 2018.

\bibitem[Zhu et~al.(2022)Zhu, Peng, Larsson, Xu, Bao, Cui, Oswald, and Pollefeys]{zhu2022nice}
Zihan Zhu, Songyou Peng, Viktor Larsson, Weiwei Xu, Hujun Bao, Zhaopeng Cui, Martin~R Oswald, and Marc Pollefeys.
\newblock Nice-slam: Neural implicit scalable encoding for slam.
\newblock In \emph{Proceedings of the IEEE/CVF Conference on Computer Vision and Pattern Recognition}, pages 12786--12796, 2022.

\end{thebibliography}
}

% \clearpage
% % \setcounter{page}{1}
% % \maketitlesupplementary
% % \appendix
% % \section*{APPENDIX: \uppercase{Learning Generalizable Next-Best-View Policies for Embodied 3D Scene Exploration}}

% % \appendix
% % \twocolumnbreak  % Forces a break before switching
% % \section*{\centering APPENDIX: \uppercase{Your Paper Title}}
% \appendix
% \onecolumn  % Switch to one column for the appendix title
% \begin{center}
%     \section*{Learning Generalizable Next-Best-View Policies for Embodied 3D Scene Exploration}
%     \section*{Supplementary Material}
% \end{center}

% \twocolumn  % Switch back to two columns for appendix content
% \section{Implementation Details}
% \subsection{Occupancy Grid Mapping Algorithm}
% \label{app:grid}

\clearpage
\maketitlesupplementary
\appendix

\section{Datasets}
\label{app:data}

    \begin{table*}[!b]
    \small
    \center
    \vspace{-3pt}
    \setlength\tabcolsep{15pt}  % width of column
    \renewcommand{\arraystretch}{1.2}   % height of row
    \captionof{table}{The sources of our collected and processed scene meshes. We created the meshes of complex scenes from ProcTHOR-10K using our export script. After manually filtering out the meshes of complex indoor scenes from \textbf{ProcTHOR-10K}, \textbf{HSSD}, \textbf{Gibson}, and \textbf{MP3D}, we preprocess them and split them into \textbf{1024 training scenes} and \textbf{128 test scenes}.}
    \begin{tabular}{|c|c|c|c|c|c|}
    \hline
    \textbf{Dataset} & \textbf{Total Amount} & \textbf{Type / Mode} & \textbf{Amount} & \textbf{Training} & \textbf{Test} \\ \hline
    \multirow{7}{*}{ProcTHOR-10K} & \multirow{7}{*}{\begin{tabular}[c]{@{}c@{}}896 (train)\\ \\ 76 (test)\end{tabular}} & 4-room & 284 & 256 & 28 \\ \cline{3-6} 
     &  & 5-room & 164 & 164 & 0 \\ \cline{3-6} 
     &  & 2-bed-2-bath & 280 & 256 & 24 \\ \cline{3-6} 
     &  & 7-room-3-bed & 114 & 96 & 18 \\ \cline{3-6} 
     &  & 8-room-3-bed & 28 & 28 & 0 \\ \cline{3-6} 
     &  & 12-room & 64 & 64 & 0 \\ \cline{3-6} 
     &  & 12-room-3-bed & 38 & 32 & 6 \\ \hline
    \multirow{3}{*}{HSSD} & \multirow{3}{*}{32 (train), 10 (test)} & easy & 10 & \multirow{3}{*}{32} & \multirow{3}{*}{10} \\ \cline{3-4}
     &  & medium & 8 &  &  \\ \cline{3-4}
     &  & hard & 24 &  &  \\ \hline
    Gibson & 96 (train), 24 (test) & real-scan & 120 & 96 & 24 \\ \hline
    MP3D & 18 (test) & real-scan & 18 & 0 & 18 \\ \hline
    \textbf{Total (Ours)} & \textbf{1152} & \textbf{mixed} & \textbf{1152} & \textbf{1024} & \textbf{128} \\ \hline
    \end{tabular}
    \label{app:table_data}
    \end{table*}

\subsection{Data Creation \& Data Filtering}
    There are several well-known datasets of indoor scenes, including ScanNet~\cite{dai2017scannet}, Replica~\cite{straub2019replica}, Gibson~\cite{xia2018gibson}, and Matterport3D (MP3D)~\cite{chang2017matterport3d}. However, the embodied AI community faces several challenges when working with these datasets: 1) there is a limited number of high-quality real-scan datasets, where ``high-quality" refers to watertight surfaces, well-organized layouts, and unified reconstruction quality; 2) synthetic scenes often lack realistic features such as collision-rich layouts and high-fidelity furniture assets; 3) public datasets of indoor scenes employ different organizational structures, making it difficult to collect scene meshes with unified formats and scales, which impedes their effective use in simulation and policy learning.

    In this work, we collect and preprocess 1152 high-quality meshes of complex indoor scenes from ProcTHOR-10K~\cite{deitke2022procthor}, HSSD~\cite{khanna2024hssd}, Gibson, and MP3D to learn a generalizable exploration policy for active 3D mapping. To leverage the digital assets in the general format originally coupling in AI2THOR~\cite{kolve2022ai2thor} and Unity~\cite{juliani2020unity} backend, we created 972 meshes of complex scenes from ProcTHOR-10K using our export script.  After manually filtering out the high-quality meshes of complex indoor scenes from these datasets, we preprocess them into a unified format and scale. Finally, these meshes are split into 1024 training scenes and 128 test scenes in our benchmark. The details of the data source are shown in Table~\ref{app:table_data}. The details of preprocessing each dataset are as follows:

    \noindent \textbf{ProcTHOR-10K.}
    ProcTHOR~\cite{deitke2022procthor} and AI2-THOR~\cite{kolve2022ai2thor} have empowered the research community to procedurally generate fully interactive, high-fidelity indoor scenes with diverse layouts for robotic training at scale. They introduce the ProcTHOR-10K dataset as templates of generated layouts, which includes 10,000 diverse indoor scenes. However, the original AI2-THOR platform and ProcTHOR dataset are limited by their reliance on the Habitat platform~\cite{savva2019habitat} and Unity Engine~\cite{juliani2020unity} for asset simulation and management, which constrains the extensibility of these valuable digital assets. To address this limitation, we developed an autonomous script that batch exports the generated scenes from Unity Editor to mesh files. This approach enables the procedural generation of scene meshes with editable content, including materials, floorplans, object placement, and controllable connectivity. Notably, users can generate an arbitrary number of scenes and then export them into mesh files using our exporting script. These infinitely generated 3D assets can be utilized for both policy training and digital content creation for AR/VR. We will release both the export script and the created assets.

    \noindent \textbf{Habitat Synthetic Scenes Dataset (HSSD).} The HSSD dataset comprises 211 meticulously crafted 3D environments specifically designed to facilitate generalization capabilities within realistic 3D environments. This collection is characterized by its professionally curated digital assets and intricate spatial arrangements. We have selected 42 exemplary indoor scenes from this dataset, which serve as valuable photorealistic synthetic sources for exploration tasks. While decorative elements improve scene realism, they are unnecessary for policy learning and introduce excessive computational costs that hinder large-scale simulation training. Consequently, the scenes underwent systematic preprocessing through geometric simplification, particularly focusing on decorative elements and doors that might impede cross-room navigability.
    
    \noindent \textbf{Gibson \& Matterport3D.} Gibson and Matterport3D are public real-scan datasets providing hundreds of complex indoor scenes. However, the styles of these scene meshes exhibit significant variation, and the mesh quality is too inconsistent for direct use. Therefore, we filter these two datasets by the following criteria: 1) accurate reconstruction with minimal floaters and artifacts, 2) enclosed scene mesh with a nearly watertight external surface, and 3) one-floor structure. As a result, we obtain $120$ diverse high-quality scene meshes from Gibson for training and evaluation. Also, we split all $18$ selected meshes from MP3D for cross-dataset and out-of-domain evaluation.

\subsection{Data Preprocessing}
    \noindent \textbf{Mesh Preprocessing.} 
    To standardize the coordinate systems across scene meshes from different datasets, we transform all meshes such that their origin points lie at the geometric center of the floors, with the height direction isotropic to the +Z-axis. The transformation scripts are implemented in Python using Open3D library~\cite{Zhou2018}

    \noindent \textbf{Ground-Truth Point Cloud.} 
    We generate ground-truth point clouds using the Poisson Disk sampling method~\cite{yuksel2015sample}, implemented in Open3D~\cite{Zhou2018}, to sample 100,000 points from the 3D scene meshes. To simplify visibility determination, we voxelize these point clouds at a specified resolution (grid size = 128 in this work) and filter out obviously invisible points, such as internal points enclosed within surfaces. These voxelized points serve as the ground-truth point clouds for the meshes and are used to compute key metrics like coverage ratio.

\subsection{Dataset Split \& Training Stages}
    Due to memory constraints and computational efficiency, we distributed the 1,024 training scenarios across two sequential training stages (i.e., stage 1 \& stage 2). The final checkpoint from the first training stage served as parameter initialization for the subsequent stage. The one-stage exploration policy is optimized through 2.5k iterations and uses approximately 48 hours of training time on a single GeForce RTX 4090 GPU. 

    The dataset split of the two training stages is as follows. \textbf{Stage 1}: ``procthor-4-room (256)", ``procthor-5-room (164)", ``procthor-8-room-3-bed (28)", ``procthor-12-room-3-bed (32)", ``hssd (32)". \textbf{Stage 2}: ``procthor-2-bed-2-bath (256)", ``procthor-7-room-3-bed (96)", ``procthor-12-room (64)", ``gibson (96)".

\section{Implementation Details}
\label{app:details}
\subsection{Occupancy Grid Mapping Algorithm}
\label{app:grid}
    The goal of an occupancy mapping algorithm~\cite{thrun2002probabilistic} is to estimate the posterior probability of occupancy over voxels given the current probabilistic grid and the novel measurement event of camera ray casting. In particular, the more frequently a voxel is passed through by camera rays, the more confident the agent regards it as navigable free space.

    \noindent \textbf{PyCUDA-based Bresenham's Line Algorithm.} Before updating the probabilistic occupancy grid, Bresenham's line algorithm~\cite{bresenham1965algorithm} is implemented to cast the ray path in 3D space between the camera viewpoint and the endpoints among the point cloud back-projected from captured depth maps. To accelerate the computing efficiency, we use PyCUDA~\cite{klockner2012pycuda} to implement Bresenham's line algorithm.

    \noindent \textbf{Derivation of Map Updating.} In practice, we adhere to the algorithm implementation outlined in GenNBV~\cite{chen2024gennbv}. A comprehensive explanation of the methodology, along with the experimental results, is provided in the appendix of GenNBV. We provide the key derivation of the log-odds formulation of occupancy probability as follows:

     Before updating the probabilistic occupancy map $G_t$, Bresenham's line algorithm is implemented to cast the ray path in 3D space between the camera viewpoint and the endpoints among the point cloud back-projected from $D_{t+1}$. % We then update the occupancy probability of the voxels passed by camera rays following the Bayesian principles. 
     According to the classical occupancy grid mapping algorithm~\citep{thrun2002probabilistic}, we have the log-odds formulation of occupancy probability:
    \begin{equation}
        \label{eqn:OGM}
        \log Odd(m_i|z_j) = \log Odd(m_i) + \log\frac{p(z_j|m_i=1)}{p(z_j|m_i=0)},
    \end{equation}
    where $m_i$ denotes the occupancy probability of $i^{th}$ voxel in the map $G_t$, $z_j$ is the measurement event that $j^{th}$ camera ray passes through this voxel. 

    For the item $C=\log\frac{p(z_j|m_i=1)}{p(z_j|m_i=0)}$, there are only two cases for the measurement event in fact: $z_j=0$ or $z_j=1$. Thus, if the measurement event $z_j$ (i.g., the voxel is passed through by the $j^{th}$ camera ray) happens, we’ll update the occupancy by adding the value of $C_1=\log\frac{p(z_j = 1 | m_i = 1)}{p(z_j = 1 | m_i = 0)}$. If it’s not passed, we’ll add the value $C_2 = \log\frac{p(z_j = 0 | m_i = 1)}{p(z_j = 0 | m_i = 0)}$. The values of $C_1$ and $C_2$ can be set as empirical constants, depending on factors such as the accuracy of ray casting and the confidence of each ray. Actually, $C’ = |\frac{C_1}{C_2}|$. We set a high value for $C'$ (i.e., high confidence) because our experiments are based on the realistic simulator and accurate observations like depth maps.

    Therefore, we can update the occupancy status of each voxel in the map $G_t$ by adding a constant for each ray casting process. Note that the probabilistic occupancy map $F^G$ is continuously updated within an episode. Finally, the occupancy status of voxels can be classified into three categories: unknown, occupied, and free, by setting an empirical threshold.

\subsection{A* Path-Finding Algorithm}
\label{app:astar}
    \noindent To evaluate the navigability between the agent's current position and predicted 3D target position, we implement a classic A* path-finding algorithm~\cite{hart1968formal} in 3D space. We developed a CUDA-based implementation of the algorithm, increasing the computational efficiency. The system classifies a target position as unnavigable if the computed path length exceeds a predefined threshold, ensuring that our NBV policy predicts reliable and safe target poses.

    Most previous works regard path-finding algorithms as a local policy and define a few movement commands (e.g., move forward $10cm$, turn left $30\degree$) as their action space. However, we don't follow this paradigm in our work for the following main reasons: 1) The key challenge of exploration policy is to determine the next best viewpoint, instead of the next neighbor step. Classic and learning-based planning and control methods both are capable of handling the control process toward the target viewpoint. 2) Popular action space, which consists of {move forward $10cm$, turn left $30\degree$, turn right $30\degree$}, makes redundant waypoints that produce inefficient trajectory, non-smooth control, and costly frequency of map updating, planning, and control.

% \subsection{Collision Detection}
% \label{app:collision}
% We classify the following three cases into collision.

\subsection{Key Hyperparameters and Details}
\label{app:param}

    \begin{table}[tbp!]
        \centering
        \caption{The key hyperparameters for our policy learning.}
        \begin{tabular}{ll}
        \toprule 
        \textbf{Term} & \textbf{Value} \\
        \midrule 
        Optimizer & Adam \\
        Optimization batch size & 128 \\
        Learning rate & 0.0001 \\
        Training Iterations & 2500 \\
        % Max Episode length \todo & 100 \\
        Training Environments & 32 \\
        N steps & 512 \\
        N epochs & 4 \\
        Buffer size & 30 \\
        Value coefficient & 0.8  \\ 
        Entropy coefficient & 0.01  \\ 
        Discount factor $\gamma$ & 0.99  \\ 
        GAE $\tau$ & 0.99  \\ 
        PPO clipping & 0.2  \\ 
        \bottomrule
        \end{tabular}
        \label{table:app_params}
    \end{table}

    The key hyperparameters of our policy learning are shown in Table~\ref{table:app_params}. Our implementation builds upon the codebase of Legged Gym~\cite{rudinlearning} and utilizes the PPO implementation from Stable-Baselines3~\cite{raffin2021stable}, which is developed in PyTorch~\cite{paszke2019pytorch}. 

    \noindent\textbf{PPO Implementation.} Specifically, given our parameterized policy $\pi_\theta$, the objective of PPO is to maximize the following function:
        \begin{equation}
        L(\theta) = \mathbb{E}_{t}\left[\frac{\pi_\theta(a_t|s_t)}{\pi_{\theta_\text{old}}(a_t|s_t)} A^{\pi_{\theta_\text{old}}}(s_t, a_t)\right],
        \end{equation}
     where $A^{\pi_{\theta_\text{old}}}(s_t, a_t)$ is the advantage function that measures the value of taking action $a_t$ at state $s_t$ under the current policy $\pi_{\theta_\text{old}}$.
    To prevent significant deviation of the new policy from the old policy, PPO incorporates a clipped surrogate objective function:
        \begin{equation}
            \begin{split}
            L^\text{CLIP}(\theta) =  &\mathbb{E}_{t}[\text{min}(\eta_t(\theta) A^{\pi_{\theta_\text{old}}}(s_t, a_t), \\
                                    &\text{clip}(r_t(\theta), 1 - \epsilon, 1 + \epsilon) A^{\pi_{\theta_\text{old}}}(s_t, a_t))] ,
            \end{split}
        \end{equation}
        where $\eta_t(\theta) = \frac{\pi_\theta(a_t|s_t)}{\pi_{\theta_\text{old}}(a_t|s_t)}$ and $\epsilon$ is a hyper-parameter that controls the size of the trust region.  

    \noindent\textbf{Onboard Cameras.} We assume that upon reaching the target pose, the agent performs four sequential 90-degree rotations and captures an observation at each orientation. To mitigate the significant computational overhead associated with repeated rendering during rotation, we implemented a simulation of four cameras mounted on the agent, with their headings oriented at 90-degree intervals.

    \noindent\textbf{Keyframe Budget During Inference.} The budget $T=50$ during inference was set based on the average exploration keyframes $\overline{T}=31.78$ across methods. This value balances policy completeness and computational efficiency while not compromising the generalizability.

\subsection{Implementation of Baseline Methods}
We implement and evaluate the following works in our benchmark to demonstrate the superiority of our proposed method: 1) \textbf{Random Policy} randomly samples actions from Gaussian distribution within the action space. 2) \textbf{Vacuum} simulates a heuristic exploration policy for robot vacuums. We follow~\cite{chenlearning} to let policy move straight when safe and execute a random number of 9\degree turns when a collision occurs. 3) \textbf{FBE} always moves the agent towards the navigable nearest boundary between observed and unknown areas. 4) \textbf{UPEN~\cite{georgakis2022uncertainty}} estimates the information gain of candidate trajectories sampled by RRT~\cite{lavalle1998rapidly}, where the gain is estimated by model ensembles. We reduce the number of ensembling models and the number of layers due to the limited memory. 5) \textbf{ANM~\cite{yan2023anm}} learns exploration in a neural implicit representation optimization framework. It estimates the information gain of candidate poses by three empirical criteria. We replace its RL-based local planner with our A-star planner. 6) \textbf{ANS~\cite{chaplotlearning}}: The original implementation of this policy relies on a global normalized map with unified resolution as input instead of an egocentric observed map, thus it cannot be directly generalized to unknown environments. We adapt this policy to a generalizable pipeline that takes a global egocentric map as input and also augment the policy learning with our random initialization strategy to make it generalizable. Given ground-truth poses, we adapt the original active SLAM system to an active mapping system.
7) \textbf{OccAnt~\cite{ramakrishnan2020occupancy}}: Similar to ANS, we also provide ground-truth poses to adapt it to an active mapping system. Due to the limited storage, we reduced its map resolution and consequently increased the voxel size to ensure a similar perceptual range.

\subsection{Visualization Implementation}
All trajectories in Figure~\ref{fig:teaser}, \ref{fig:motivation}, \ref{fig:vis} are actual results. We record waypoints/keyframes and reconstruct the active mapping process using Open3D for offline visualization.

\section{Training Strategy}
\subsection{Scene Updating Strategy}
\label{app:sceneupdate}
    To enhance the generalizability, we create a training set including $512$ diverse indoor scenes in each training stage from our \oursbench. However, we cannot launch such a large number of parallel training environments in simulation due to the limitations of computational efficiency and memory. As introduced in Sec.~\ref{sec:method_strategy}, we adopt a workaround to update the active scene in the limited training environments. We launch 32 training environments in Isaac Gym, and load $16$ different scenes as a sampling set for each environment. During training, there is a predefined probability of $p$ to randomly activate a scene in each environment's inactive sampling set. In particular, we move the replaced scenes to the inactive area (i.e., out of the agents' movement space) in the simulator and move the sampled scenes to the active areas of corresponding environments. As shown in Table~\ref{tab:ablation_design}, we found that frequently updating the active scenes utilizes the diversity of training scenes, and improves the generalizability of policies.

\subsection{Capturing at Long-Term Target Positions}
    Previous work~\cite{chenlearning} typically employs discrete single-step actions, such as moving forward $10cm$ or turning left $10$. However, this single-step planning and control approach is inconsistent with real-world robotic systems and significantly increases the computational cost of simulation for RL-based policy training. Moreover, real-world inference of this setup requires numerous policy network iterations, making it prohibitively time-consuming. Therefore, we optimize our approach to predict navigable next-best viewpoints in free space rather than relying on classical single-step actions.

    To enhance practical effectiveness, we capture four surrounding views at each predicted position, simulating the scanning process. This multi-view setup provides a broader spatial context and enables more effective long-term planning during policy training.

\section{Additional Results}
\subsection{The Reward of Trajectory Efficiency}
    \begin{table}[!h]
    \centering
    \footnotesize
    \setlength{\tabcolsep}{3.5pt}
    \renewcommand{\arraystretch}{1.1}
    \setlength{\belowcaptionskip}{-10pt}
    \captionof{table}{The effect of path efficiency reward. $\dagger$: trained on 128 scenes and half-standard 2.5k iterations.}
    \vspace{-8pt}
    \label{tab:efficiency}
        \begin{tabular}{l|ccccc}
        \hline
            \textbf{Settings} & \textbf{Cov.} & \textbf{AUC} & \textbf{Comp.} & \textbf{KF} & \textbf{TL} \\ \hline
            \ours$\dagger$ & 60.23\% & 51.69\% & 0.89m & 23.20 & 47.32m \\ 
            \ours$\dagger$ with effi. rew. & 56.61\% & 47.91\% & 0.96m & 17.91 & 34.41m \\ 
        \hline
        \end{tabular}
    \vspace{-9pt}
    \end{table}
As shown in Table~\ref{tab:efficiency}, while implementing a generic efficiency reward term $r_t^{\mathrm{Effi}} = -1$~\cite{chen2024gennbv} indeed reduce the number of keyframes ($-5.29$) and trajectory length ($-12.91$m), it penalizes exploratory actions like detouring around obstacles, leading to conservative policies ($-3.62\%$ Cov.).

\end{document}